\documentstyle[a4]{article}
\title{Appropriate nouns with obligatory modifiers}
\author{\'Eric Laporte\thanks{Institut Gaspard-Monge, Universit\'e de Marne-la-Vall\'ee, 2, rue de la Butte-Verte, F-93166 Noisy-le-Grand CEDEX, France.}}
\date{October 22nd, 1994}
\begin{document}
\maketitle
\newcounter{numero}

\section{Introduction}

This study\footnote{I cheerfully thank Morris Salkoff for his valuable remarks on a preliminary version of this paper and for his encouragements.} is about a set of appropriate nouns {\it Napp\/} in French. A sequence is said to be appropriate to a given context \cite[1976, pp.~113-114]{Harris70} if it has the highest plausibility of occurrence in that context, and can therefore be reduced to zero.
In French, the notion of appropriateness is often connected with a metonymical restructuration of
the subject \cite{GL81}:
\begin{tabbing}(000)\=(D) =\=?*\=\kill
(\refstepcounter{numero}\arabic{numero}\label{0-5})
\>\>\>{\it Cette voiture n'est pas donn\a'ee} \\
\>=\>\>{\it Le prix de cette voiture n'est pas donn\a'e}\end{tabbing}
or of the direct object \cite[table 32R1]{BGL76}:
\begin{tabbing}(000)\=(D) =\=?*\=\kill
(\refstepcounter{numero}\arabic{numero}\label{0-6})
\>\>\>{\it La fatigue ralentit Luc} \\
\>=\>\>{\it La fatigue ralentit les mouvements de Luc}\end{tabbing}
In these examples, {\it Napp\/} =: {\it prix\/} is appropriate to (\ref{0-5}) and {\it mouvements\/} is appropriate to (\ref{0-6}).
The metonymical relation between e.g. {\it Napp\/} =: {\it prix\/} and {\it voiture,} or between {\it mouvements\/} and {\it Luc,} can usually be studied in the framework of a support verb ({\it Vsup\/}) construction \cite{Gross81} where {\it Napp\/} is the predicate:
\begin{tabbing}(000)\=(D) =\=?*\=\kill
\>\>\>{\it Cette voiture a un prix} \\
\>\>\>{\it Luc fait des mouvements}\end{tabbing}
In some sentences of the form {\it Det $N_0$ Vsup Det N Adj,} the support verb {\it Vsup\/} and the noun {\it N\/} can be considered as appropriate to their context and replaced by {\it \^etre,} leading to {\it Det $N_0$ \^etre Adj,} without any loss of information:
\begin{tabbing}(000)\=(D) =\=?*\=\kill
(\refstepcounter{numero}\arabic{numero}\label{0-7})
\>\>\>{\it Cette salle a une acoustique r\a'everb\a'erante}\\
(\refstepcounter{numero}\arabic{numero}\label{0-8})
\>\>\>{\it Cette salle est r\a'everb\a'erante}\end{tabbing}
In general, but not always, this phenomenon has a connection with a restructuration of the subject of the adjectival predicate \cite{Molinier1988}:
\begin{tabbing}(000)\=(D) =\=?*\=\kill
\>\>\>{\it Cette salle est r\a'everb\a'erante}\\ 
\>=\>\>{\it L'acoustique de cette salle est r\a'everb\a'erante}\end{tabbing} 
The appropriateness depends on both the subject and the adjective: the noun {\it couleur\/} "colour" is appropriate to:
\begin{tabbing}(000)\=(D) =\=?*\=\kill
(\refstepcounter{numero}\arabic{numero}\label{0-1})
\>\>\>{\it Ce sac est\/} ({\it brun\/} + {\it roux\/} + {\it noir\/} + {\it blanc\/})\end{tabbing}
which means:
\begin{tabbing}(000)\=(D) =\=?*\=\kill
(\refstepcounter{numero}\arabic{numero}\label{0-2})
\>\>\>{\it Ce sac a une couleur\/} ({\it brune\/} + {\it rousse\/} + {\it noire\/} + {\it blanche\/})
\end{tabbing}
but the noun {\it cheveux\/} "hair" is appropriate to:
\begin{tabbing}(000)\=(D) =\=?*\=\kill
\>\>\>{\it Luc est\/} ({\it brun\/} + {\it roux\/})\end{tabbing}
which is interpreted as:
\begin{tabbing}(000)\=(D) =\=?*\=\kill
\>\>\>{\it Luc a des cheveux\/} ({\it bruns\/} + {\it roux\/})\end{tabbing}
and normally not as:
\begin{tabbing}(000)\=(D) =\=?*\=\kill
\>\>\>{\it Luc a\/} ({\it une couleur\/} + {\it la peau\/}) ({\it brune\/} + {\it rousse\/})\end{tabbing}
On the other hand, {\it peau\/} "skin" is appropriate to:
\begin{tabbing}(000)\=(D) =\=?*\=\kill
\>\>\>{\it Luc est\/} ({\it noir\/} + {\it blanc\/})\end{tabbing}
which is interpreted as:
\begin{tabbing}(000)\=(D) =\=?*\=\kill
\>\>\>{\it Luc a la peau\/} ({\it noire\/} + {\it blanche\/})\end{tabbing}
and normally not as:
\begin{tabbing}(000)\=(D) =\=?*\=\kill
\>\>\>{\it Luc a\/} ({\it une couleur\/} + {\it des cheveux\/}) ({\it noirs\/} + {\it blancs\/})\end{tabbing}
Conversely, when a noun occurs in a construction like (\ref{0-7}), it may be appropriate to its context or not. For instance, {\it progression\/} "progression" can be removed without any loss of information from:
\begin{tabbing}(000)\=(D) =\=?*\=\kill
\>\>\>{\it La progression du convoi est rapide}\end{tabbing}
which means the same as:
\begin{tabbing}(000)\=(D) =\=?*\=\kill
\>\>\>{\it Le convoi est rapide}\end{tabbing}
On the other hand, {\it vitesse\/} "speed" is not appropriate to its context in:
\begin{tabbing}(000)\=(D) =\=?*\=\kill
\>\>\>{\it La vitesse du convoi est \a'elev\a'ee}\end{tabbing}
It can be removed, but the meaning of the resulting sentence is completely different:
\begin{tabbing}(000)\=(D) =\=?*\=\kill
\>\>?\>{\it Le convoi est \a'elev\a'e}\end{tabbing}

The notion of highest plausibility of occurrence of a term in a given context must not be
understood in a probabilistic sense. In particular, such a plausibility cannot be evaluated
through the statistic analysis of a corpus of texts. The principal cooccurrent of a linguistic
context can rather be identified through the intuition of a paraphrastic relation between
sentences like (\ref{0-1}) and (\ref{0-2}). Consequently, a given linguistic context may happen to
have several principal cooccurrents, with the same meaning or with different meanings. For instance,
\begin{tabbing}(000)\=(D) =\=?*\=\kill
\>\>\>{\it Luc est camus}\end{tabbing} 
is equivalent to the following sentences, which have exactly the same meaning and roughly the same plausibility of occurrence:
\begin{tabbing}(000)\=(D) =\=?*\=\kill
\>\>\>{\it Luc a un nez camus}\\ 
\>\>\>{\it Luc a un visage camus}\end{tabbing} 
The following sentence:
\begin{tabbing}(000)\=(D) =\=?*\=\kill
\>\>\>{\it Luc est louche}\end{tabbing} 
can be interpreted, depending on its extra-linguistic context, as:
\begin{tabbing}(000)\=(D) =\=?*\=\kill
\>\>\>{\it Luc a des agissements louches}\\ 
\>\>\>{\it Luc a une allure louche}\\ 
\>\>\>{\it Luc a un comportement louche}\\ 
\>\>\>{\it Luc a une conduite louche}\end{tabbing}
Since the identification of appropriate sequences is a matter of intuition, we consider it requires a careful analysis and the use of formal criteria \cite{Molinier1988}.

In the following we examine the relations between pairs of sentences like (\ref{0-7})--(\ref{0-8}) or (\ref{0-1})--(\ref{0-2}).
In these constructions, the noun cannot be used with the determiner {\it un\/} and no modifier:
\begin{tabbing}(000)\=(D) =\=?*\=\kill
\>\>*\>{\it Cette salle a une acoustique}\\
\>\>?*\>{\it Ce sac a une couleur}\\
\>\>*\>{\it Cette sauce a une saveur}\end{tabbing}
This interdiction is not explained satisfactorily by the fact that such sentences would convey too little information.
Some sentences of the same form are accepted, though not informative at all:
\begin{tabbing}(000)\=(D) =\=?*\=\kill
\>\>\>{\it Cet ensemble a un cardinal}\end{tabbing}
or forbidden, even though they would be informative:
\begin{tabbing}(000)\=(D) =\=?*\=\kill
\>\>*\>{\it Luc a tenu des propos \a`a L\a'ea}\end{tabbing}
We will say that these constructions have an obligatory modifier, i.e. the {\it Vsup\/} construction must include a relative clause:
\begin{tabbing}(000)\=(D) =\=?*\=\kill
\>\>\>{\it Cette sauce a une saveur qui \a'etonne tout le monde}\end{tabbing}
or a substitute for it, e.g. an adjective or a prepositional complement:
\begin{tabbing}(000)\=(D) =\=?*\=\kill
\>\>\>{\it Cette sauce a une saveur\/} ({\it d\a'elicieuse\/} + {\it de viande fum\a'ee\/})\end{tabbing}
a definite determiner with a referential interpretation:
\begin{tabbing}(000)\=(D) =\=?*\=\kill
\>\>?\>{\it La sauce a cette saveur}\end{tabbing}
the determiner {\it un certain\/}:
\begin{tabbing}(000)\=(D) =\=?*\=\kill
\>\>?\>{\it Cette sauce a une certaine saveur}\end{tabbing}
or even the determiner {\it de le\/} with an intensive meaning:
\begin{tabbing}(000)\=(D) =\=?*\=\kill
\>\>\>{\it Cette sauce a de la saveur}\end{tabbing}
We will not consider predicative nouns which are appropriate to
adjectival sentences, but which can occur without any modifier:
\begin{tabbing}(000)\=(D) =\=?*\=\kill
\>\>\>{\it Ce proc\a'ed\a'e a un avenir}\\
\>\>\>{\it Luc a un look}\\
\>\>\>{\it Ce vin a un mill\a'esime}\\
\>\>\>{\it Luc a une nationalit\a'e}\\
\>\>\>{\it Luc a une religion}\\
\>\>\>{\it Ce mot a un sens}\end{tabbing}

\section{Classification}
The definitions above delimit a set of nouns of which we listed about 300. We constituted this list partly by systematic searches in conventional dictionaries, and partly by selecting entries from the lists of  \cite{Giry93}, [J.~Giry-Schneider, 1978a], \cite{Meunier81}, \cite{Vives84}, \cite{Giry87}, \cite{Gross89}. It is not exhaustive, but offers a variety of syntactic behaviours. We sorted this corpus into two classes. In the first, more numerous, class, the nouns fit into the construction {\it Det Napp de Det $N_0$ \^etre Adj\/}:
\begin{tabbing}(000)\=(D) =\=?*\=\kill
\>\>\>{\it L'acoustique de cette salle est r\a'everb\a'erante}\end{tabbing}
The other class is illustrated e.g. by {\it citoyennet\'e\/} in:
\begin{tabbing}(000)\=(D) =\=?*\=\kill
\>\>\>{\it Luc est danois}\\
\>=\>\>{\it Luc a la citoyennet\a'e danoise}\end{tabbing}
In the case of these nouns, the structure {\it Det $N_0$ Vsup Det Napp Adj\/} is the source of
\begin{itemize}
\item {\it Det $N_0$ \^etre Adj,} which gives to {\it Vsup Det Napp\/} some features of a support verb;
\item a noun phrase {\it Det Napp Adj de Det $N_0$\/} through the reduction of {\it Vsup.}
\end{itemize}

\subsection{Nouns of parts of the body}
Among the set of 300 nouns, a small subset have a special status since it is not clear whether they have an obligatory modifier in the {\it Vsup\/} construction or not. They denote parts of the body or parts of things ({\it Npb\/}). Consider, for example, the noun {\it visage\/} "face". The status of sentences like:
\begin{tabbing}(000)\=(D) =\=?*\=\kill
(\refstepcounter{numero}\arabic{numero}\label{1})
\>\>?*\>{\it Luc a un visage}\end{tabbing}
has been much discussed. Intuitively, they are so little informative that they are nearly unacceptable --- and this is why we have included them in the corpus. However, the acceptability of sentences like:
\begin{tabbing}(000)\=(D) =\=?*\=\kill
\>\>\>{\it Tout le monde a un estomac}\\
\>\>\>{\it Luc n'a pas de mains}\\
\>\>\>{\it Toute page a deux faces}\\
\>\>\>{\it Les hommes aussi ont une peau\/} (a publicity slogan)\end{tabbing}
leads one to consider (\ref{1}) and analogous sentences as --- at least theoretically --- acceptable elementary sentences. They describe the state of the world. Consequently, support verb constructions with such nouns have no obligatory modifier. Let us examine them anyway. The noun {\it visage\/} is appropriate to several adjectival sentences:
\begin{tabbing}(000)\=(D) =\=?*\=\kill
(\refstepcounter{numero}\arabic{numero}\label{2})
\>\>\>{\it Luc est\/} ({\it camus\/} + {\it glabre\/} + {\it imberbe\/} + {\it poupin\/})\end{tabbing}
are interpreted as:
\begin{tabbing}(000)\=(D) =\=?*\=\kill
(\refstepcounter{numero}\arabic{numero}\label{3})
\>\>\>{\it Luc a un visage\/} ({\it camus\/} + {\it glabre\/} + {\it imberbe\/} + {\it poupin\/})\end{tabbing}
If (\ref{1}) is accepted as a sentence, it can be used to derive (\ref{2}) and (\ref{3}) through syntactic relations. The following sequence:
\begin{tabbing}(000)\=(D) =\=?*\=\kill
(\refstepcounter{numero}\arabic{numero}\label{4})
\>\>?*\>{\it Luc a un visage, ce visage est imberbe}\end{tabbing}
is accepted as a discourse made of two elementary sentences. A relative clause can be formed by embedding the second sentence into the first in the usual way \cite{Kuroda68}:
\begin{tabbing}(000)\=(D) =\=?*\=\kill
\>\>\>{\it Luc a un visage qui est imberbe}\end{tabbing}
and [{\it qui \^etre\/} z.] leads to:
\begin{tabbing}(000)\=(D) =\=?*\=\kill
(\ref{3})
\>\>\>{\it Luc a un visage imberbe}\end{tabbing}
Now another relative clause can be formed by embedding the first sentence of (\ref{4}) into the second:
\begin{tabbing}(000)\=(D) =\=?*\=\kill
(\refstepcounter{numero}\arabic{numero}\label{6})
\>\>?\>{\it Le visage que Luc a est imberbe}\end{tabbing}
The degree of acceptability of (\ref{6}) is low but [Red. {\it Vsup\/}] applies to it: 
\begin{tabbing}(000)\=(D) =\=?*\=\kill
(\refstepcounter{numero}\arabic{numero}\label{7})
\>\>\>{\it Le visage de Luc est imberbe}\end{tabbing}
and a relation of restructuration leads to:
\begin{tabbing}(000)\=(D) =\=?*\=\kill
(\refstepcounter{numero}\arabic{numero}\label{8})
\>\>\>{\it Luc est imberbe\/} (E + ?{\it de visage\/})\end{tabbing}
This analysis is shown in Fig.~\ref{avoirNpc}.
\begin{figure}
\begin{tabbing}
(D) \=N0 avoir\=(D) \=un Npb qui \^etre Adj\=(D) \=\kill
\>\>(A)\>{\it $N_0$ avoir un Npb, ce Npb \^etre Adj}\\
\>\>Rel.\>\>Rel.\\
(B)\>{\it $N_0$ avoir un Npb qui \^etre Adj}\>\>\>(D)\>{\it Npb que $N_0$ avoir \^etre Adj}\\
\>[{\it qui \^etre\/} z.]\>\>\>\>[Red. {\it Vsup\/}]\\
(C)\>{\it $N_0$ avoir un Npb Adj}\>\>\>(E)\>{\it Npb de $N_0$ \^etre Adj}\\
\>\>\>\>\>[Restruct.]\\
\>\>\>\>(F)\>{\it $N_0$ \^etre Adj\/} (E + {\it Prep Npb\/})
\end{tabbing}
\caption{Derivation of (9) and (13). Free determiners are omitted.}\label{avoirNpc}
\end{figure}

For other {\it Npb\/}'s, (D) is generally even less acceptable than (\ref{6}):
\begin{tabbing}(000)\=(D) =\=?*\=\kill
\>\>?*\>{\it La peau qu'a Luc est bronz\a'ee}\end{tabbing}
but we maintain it as a theoretical sentence. If we reject (\ref{6}), a direct relation between (\ref{4}) and (\ref{7}) is necessary, not only for the present construction but also in order to account for the use of other noun phrases of the form {\it Le Npb de Nhum\/} as actants in sentences. In Fig.~\ref{avoirNpc},
the construction (F) contains a prepositional complement. This complement is never obligatory;
the preposition and the determiner depend on {\it Npb\/} and on the adjective.

Fig.~\ref{listeNpc} gives 11 nouns which provide examples of the relations of Fig.~\ref{avoirNpc}. Each noun {\it Npb\/} is followed by one or more examples of adjectives to which {\it Npb\/} is appropriate, and a few distributional properties. For example, {\it visage\/} appears in the list because of sentences like
(\ref{2})--(\ref{8}). In each line, each distributional property is described by a sign, which is a {\tt +} if the entry has the
property and a  {\tt -} if it does not. The properties are the following:
\begin{enumerate}
\item {\it $N_0$\/} =: {\it Nhum,} i.e. the subject of the {\it Vsup\/} construction may be a human noun.
\item {\it $N_0$\/} =: {\it N-hum,} the subject of the {\it Vsup\/} construction may be a non-human noun.
\item {\it Npb sing,} the appropriate noun may be singular if {\it $N_0$\/} is semantically singular.
\item {\it Npb plur,} the appropriate noun may be plural if {\it $N_0$\/} is semantically singular.
\item {\it Det $N_0$ \^etre Adj Prep Det Npb.}
\end{enumerate}
These properties concern the sentences which can take all the forms in Fig.~\ref{avoirNpc}, including (E) and (F), and for which {\it Npb\/} is appropriate
to (F): for example, in the construction
\begin{tabbing}(000)\=(D) =\=?*\=\kill
\>\>\>{\it Luc avoir Det cheveu(x) brun(s),}\end{tabbing}
{\it Npb\/} =: {\it cheveu\/} may be singular, but then {\it cheveu\/} is not appropriate to {\it Luc est brun.} The codes at the end of the lines identify syntactic tables of other authors where the same {\it Vsup\/} construction of the noun is described.
\begin{figure}[htbp]
\begin{tabbing}
visage \=visage visage \=camus, glabre, imberbe, poupin \=damasquine\=\kill
\>{\it Npb}\>{\it Adj}\>{\tt 12345}\>{\sf Table}\\ \\
\>{\it bord}\>{\it dent\a'e de, sinu\a'e de}\>{\tt -++++}\>{\sf AN08}\\
\>{\it cheveu}\>{\it blond, brun, roux}\>{\tt +--++}\>{\sf APP1}\\
\>{\it corps}\>{\it svelte, perclus de}\>{\tt +-+-+}\>{\sf APP1}\\
\>{\it dedans}\>{\it spacieux}\>{\tt -++-+}\\
\>{\it dehors}\>{\it laqu\a'e de, damasquin\a'e de}\>{\tt -++-+}\>{\sf AN08}\\
\>{\it face}\>{\it vierge de}\>{\tt -++++}\>{\sf AN08}\\
\>{\it jambe}\>{\it cagneux}\>{\tt +--++}\\
\>{\it main}\>{\it expert \a`a}\>{\tt +-+++}\\
\>{\it nez}\>{\it camard, camus}\>{\tt +-+--}\\
\>{\it peau}\>{\it moite de}\>{\tt +-+-+}\\
\>{\it visage}\>{\it camus, glabre, imberbe, poupin}\>{\tt +-+-+}\end{tabbing}
\caption{List of appropriate {\it Npb\/}'s.}\label{listeNpc}
\end{figure}

\boldmath 
\subsection{Nouns with {\it Det Napp de Det $N_0$ \^etre Adj}}
\unboldmath 
Most of the nouns in the corpus can be used in the construction {\it Det Napp de Det $N_0$ \^etre Adj.} In general, in the {\it Vsup\/} construction, a modifier is obligatory but it commutes with
the determiner {\it un certain\/}:
\begin{tabbing}(000)\=(D) =\=?*\=\kill
\>\>\>{\it Cette salle a une acoustique\/} ({\it mate\/} + *E)\\
(\refstepcounter{numero}\arabic{numero}\label{10})
\>\>?\>{\it Cette salle a une certaine acoustique}
\end{tabbing}
Sentences like (\ref{10}) sound strange in isolation, but are clearly acceptable when followed by a sentence or an adverb which concerns {\it Napp\/}:
\begin{tabbing}(000)\=(D) =\=?*\=Cette salle\=\kill
\>\>\>{\it Cette salle a une certaine acoustique,}\\
\>\>\>\>({\it comment qualifier cette acoustique~?}\\
\>\>\>\>+ {\it essayez de vous souvenir}\\
\>\>\>\>+ {\it les graves passent bien}\\
\>\>\>\>+ {\it quelle qu'elle soit\/})
\end{tabbing}
In particular, consider the following discourse:
\begin{tabbing}(000)\=(D) =\=?*\=\kill
(\refstepcounter{numero}\arabic{numero}\label{11})
\>\>\>{\it Cette salle a une certaine acoustique, cette acoustique est mate}
\end{tabbing}
It is made of two sentences which are little autonomous, i.e. the first sentence is rather pointless in isolation, and in the second one the acoustics is necessarily that of some place or another. There is an analogy between (\ref{11}) and the construction (A) above, but the determiner of {\it acoustique\/} in (\ref{11}) is {\it un certain\/} whereas that of {\it Npb\/} in (A) is {\it un.} The sentences of (\ref{11}) can be considered as elementary sentences: in (\ref{10}), {\it un certain\/} is the most neutral possible substitute for the obligatory modifier, so it appears as the minimal determiner in that {\it Vsup\/} construction.

A relative clause can be formed by embedding the second sentence into the first:
\begin{tabbing}(000)\=(D) =\=?*\=\kill
\>\>\>{\it Cette salle a une acoustique qui est mate}\end{tabbing}
and [{\it qui \^etre\/} z.] leads to:
\begin{tabbing}(000)\=(D) =\=?*\=\kill
(\refstepcounter{numero}\arabic{numero}\label{13})
\>\>\>{\it Cette salle a une acoustique mate}\end{tabbing}
On the other hand, another relative clause can be formed by embedding the first sentence of (\ref{11}) into the second:
\begin{tabbing}(000)\=(D) =\=?*\=\kill
(\refstepcounter{numero}\arabic{numero}\label{14})
\>\>\>{\it L'acoustique qu'a cette salle est mate}\end{tabbing}
[Red. {\it Vsup\/}] applies: 
\begin{tabbing}(000)\=(D) =\=?*\=\kill
\>\>\>{\it L'acoustique de cette salle est mate}\end{tabbing}
and a relation of restructuration leads to:
\begin{tabbing}(000)\=(D) =\=?*\=\kill
(\refstepcounter{numero}\arabic{numero}\label{16})
\>\>\>{\it Cette salle est mate}\end{tabbing}

This analysis is shown in Fig.~\ref{avoirNapp}.
\begin{figure}
\begin{tabbing}
(D) \=N0 avoir\=(D) \=un Npb qui \^etre Adj\=(D) \=\kill
\>\>(A)\>{\it $N_0$ Vsup un certain Napp, ce Napp \^etre Adj}\\
\>\>Rel.\>\>Rel.\\
(B)\>{\it $N_0$ Vsup un Napp qui \^etre Adj}\>\>\>(D)\>{\it Napp que $N_0$ Vsup \^etre Adj}\\
\>[{\it qui \^etre\/} z.]\>\>\>\>[Red. {\it Vsup\/}]\\
(C)\>{\it $N_0$ Vsup un Napp Adj}\>\>\>(E)\>{\it Napp de $N_0$ \^etre Adj}\\
\>\>\>\>\>[Restruct.]\\
\>\>\>\>(F)\>{\it $N_0$ \^etre Adj\/} (E + {\it Prep Napp\/})\end{tabbing}
\caption{Derivation of (16) and (18). Free determiners are omitted.}\label{avoirNapp}
\end{figure}
When {\it Napp\/} is obligatorily plural, the determiner {\it certains\/} is an acceptable plural of {\it un certain\/} in Fig.~\ref{avoirNapp}:
\begin{tabbing}(000)\=(D) =\=?*\=\kill
\>\>\>{\it Luc a certains agissements, ces agissements sont louches}\\
\>\>\>{\it Luc a des agissements louches}\\
\>\>\>{\it Luc est louche\/} (E + ({\it par\/} + {\it dans\/}) {\it ses agissements\/})\end{tabbing}

In some cases, the following additional form is observed:
\begin{tabbing}(000)\=(D) =\=?*\=\kill
\>\>\>{\it Det $N_0$ Vsup quelque chose de Adj}\\
\>=:\>\>{\it Ce mot a quelque chose de p\a'ejoratif}\\
\>\>*\>{\it Cet animal a quelque chose de cavernicole}\end{tabbing}
but when this form is observed, its source is not clear.

The constructions (A), (B), (C), (D), which contain a {\it Vsup,} must be adapted to take account of {\it Vsup\/} constructions where {\it Napp\/} is not the direct object of {\it Vsup.} For example, by a {\it Vsup\/} substitution,
\begin{tabbing}(000)\=(D) =\=?*\=\kill
\>\>?\>{\it Cette salle a une certaine atmosph\a`ere}\\
\>=\>\>{\it Une certaine atmosph\a`ere r\a`egne dans cette salle}
\end{tabbing}
which leads to e.g.:
\begin{tabbing}(000)\=(D) =\=?*\=\kill
\>(C)\>\>{\it Un Napp Adj r\a'egner Loc Det $N_0$}\\
\>=:\>\>{\it Une atmosph\a`ere studieuse r\a`egne dans cette salle}\\
\>(D)\>\>{\it Det Napp qui r\a'egner Loc Det $N_0$ \^etre Adj}\\
\>=:\>\>{\it L'atmosph\a`ere qui r\a`egne dans cette salle est studieuse}\end{tabbing}
Similarly, {\it \^etre de\/} is often an acceptable variant of {\it Vsup\/} =: {\it avoir\/} [J.~Giry-Schneider, 1993, table AN07]:
\begin{tabbing}(000)\=(D) =\=?*\=\kill
\>\>\>{\it Ce chien\/} ({\it a\/} + {\it est d'\/}) {\it une certaine race}\\
\>(C)\>\>{\it Ce chien\/} ({\it a\/} + {\it est d'\/}) {\it une race b\^atarde}\\
\>(D)\>\>{\it La race\/} ({\it qu'a\/} + {\it dont est\/}) {\it ce chien est b\^atarde}
\end{tabbing}
Another variant of {\it Vsup\/} =: {\it avoir\/} leads to additional forms. For some nouns, the construction:
\begin{tabbing}(000)\=(D) =\=?*\=\kill
\>\>\>{\it Det $N_0$ avoir un certain Napp}\\
\>=:\>\>{\it Luc a un certain brio}\end{tabbing}
has a variant:
\begin{tabbing}(000)\=(D) =\=?*\=\kill
\>\>\>{\it Un certain Napp \^etre Loc Det $N_0$}\\
(\refstepcounter{numero}\arabic{numero}\label{16-1})
\>=:\>?\>{\it Un certain brio est en Luc}\end{tabbing}
which is used only in literary styles\footnote{In everyday French, a noun subject of {\it \^etre\/} generally has a definite determiner.}, but
which is the source of other constructions
\begin{itemize}
\item with a binding operator {\it avoir\/} \cite{Gross81}:
\begin{tabbing}(000)\=(D) =\=?*\=\kill
\>\>\>{\it Det $N_0$ avoir un certain Napp Loc lu$i^0$}\\
\>=:\>\>{\it Luc a un certain brio en lui,}
\end{tabbing}
hence:
\begin{tabbing}(000)\=(D) =\=?*\=\kill
\>\>\>{\it Luc a en lui un brio \a'eblouissant}\\
\>\>\>{\it Luc a en lui quelque chose d'\a'eblouissant\/};\end{tabbing}
\item with {\it il y avoir\/} \cite[tables AN07 and AN08]{Giry93}:
\begin{tabbing}(000)\=(D) =\=?*\=\kill
\>\>\>{\it Il y avoir un certain Napp Loc Det $N_0$}\\
\>=:\>\>{\it Il y a un certain brio en Luc,}\end{tabbing}
hence:
\begin{tabbing}(000)\=(D) =\=?*\=\kill
\>\>\>{\it Le brio qu'il y a en Luc est \a'eblouissant}\\
\>\>\>{\it Il y a en Luc quelque chose d'\a'eblouissant}\end{tabbing}
\end{itemize}

The degree of acceptability of (D) is sometimes lower than that of (\ref{14}):
\begin{tabbing}(000)\=(D) =\=?*\=\kill
\>\>\>{\it La condition\/} (*{\it qu'a\/} + ?{\it dont est\/}) {\it Luc est roturi\a`ere}\end{tabbing}

In fact, in the case of certain nouns, the regular syntax of Fig.~\ref{avoirNapp} is partly frozen and only a subset of the constructions
(A), (B), (C), (D), (E), (F) is acceptable.

For {\it acad\'emie, dehors, plume, sang,} used in more or less obsolete constructions with {\it Vsup\/} =: {\it avoir,}
(A) is unacceptable and {\it Napp\/} cannot appear in a separate clause:
\begin{tabbing}(000)\=(D) =\=?*\=\kill
\>\>\>{\it Les dehors\/} ({\it qu'a\/} + {\it de\/}) {\it Luc sont n\a'eglig\a'es}\\
\>\>?*\>{\it Luc a certains dehors}
\end{tabbing}
A direct relation between (D) and (B) must thus be envisaged:
\begin{tabbing}(000)\=(D) =\=?*\=\kill
(\refstepcounter{numero}\arabic{numero}\label{19})
\>\>?\>{\it Les dehors qu'a Luc sont n\a'eglig\a'es}\\
(\refstepcounter{numero}\arabic{numero}\label{20})
\>=\>?\>{\it Luc a des dehors qui sont n\a'eglig\a'es}
\end{tabbing}
This relation appears as the application of a binding operator {\it avoir\/} to (D), as in frozen sentences of table E01 \cite{Gross88}:
\begin{tabbing}(000)\=(D) =\=?*\=\kill
\>\>?\>{\it Le calme qu'a Luc est olympien}\\
\>=\>?\>{\it Luc a un calme qui est olympien}
\end{tabbing}
The differences are that in (\ref{19}) and (\ref{20}) the adjective is not frozen with {\it dehors,} and that {\it dehors\/} is appropriate to:
\begin{tabbing}(000)\=(D) =\=?*\=\kill
\>\>\>{\it Luc est n\a'eglig\a'e}\end{tabbing}
whereas *{\it Luc est olympien\/} is not accepted. In such a construction, it is not easy to assign an elementary sentence to the noun {\it Napp.} It can be seen as a "support" for the adjective in {\it Det Napp que Det $N_0$ avoir \^etre Adj,} in the same sense as a {\it Vsup\/}
is a support for a predicative noun in a {\it Vsup\/} construction. However, {\it Napp\/} can also be used as
an actant without an adjective:
\begin{tabbing}(000)\=(D) =\=?*\=\kill
\>\>?\>{\it Les dehors qu'a Luc s\a'eduisent Marie}\\
\>\>?\>{\it Marie admire les dehors qu'a Luc}
\end{tabbing}

These features are even more marked for nouns like {\it c\oe ur\/} "heart" which enter neither into (A) nor into (D):
\begin{tabbing}(000)\=(D) =\=?*\=\kill
\>(A)\>*\>{\it Luc a un certain c\oe ur, ce c\oe ur est tendre}\\
(\refstepcounter{numero}\arabic{numero}\label{23})
\>(B)\>?\>{\it Luc a un c\oe ur qui est tendre}\\
(\refstepcounter{numero}\arabic{numero}\label{23bis})
\>(C)\>\>{\it Luc a un c\oe ur tendre}\\
\>(D)\>*\>{\it Le c\oe ur que Luc a est tendre}\\
(\refstepcounter{numero}\arabic{numero}\label{24})
\>(E)\>\>{\it Le c\oe ur de Luc est tendre}\\
(\refstepcounter{numero}\arabic{numero}\label{24bis})
\>(F)\>\>{\it Luc est tendre}
\end{tabbing}
The construction (E) could be chosen as the basic one for these nouns. A direct relation with (B) \cite[p.~251]{Labelle83} can be envisaged. This relation appears as the application of the binding operator {\it avoir.}

As in (\ref{16-1}), some instances of (B) have a locative variant:
\begin{tabbing}(000)\=(D) =\=?*\=\kill
(\refstepcounter{numero}\arabic{numero}\label{25})
\>\>\>{\it Un Napp qui \^etre Adj \^etre Loc Det $N_0$}\\
\>=:\>?\>{\it Un c\oe ur qui est tendre est en Luc}
\end{tabbing}
which can be observed either in literary styles:
\begin{tabbing}(000)\=(D) =\=?*\=\kill
\>\>\>{\it Un c\oe ur plus tendre encore qu'autrefois sommeillait en lui}\end{tabbing}
or after the application of a binding operator:
\begin{tabbing}(000)\=(D) =\=?*\=\kill
\>\>?\>{\it Luc a en lui un c\oe ur qui est tendre}\\
\>\>?\>{\it Luc a en lui quelque chose de tendre}\end{tabbing}

The nouns {\it aspect\/} "aspect", {\it c\^ot\'e\/} "side", {\it partie\/} "part", {\it phase\/} "phase", {\it trait\/} "feature"
have the same constructions as {\it c\oe ur\/} in (\ref{23})--(\ref{24bis}), but with an indefinite
determiner in (E):
\begin{tabbing}(000)\=(D) =\=?*\=\kill
\>\>?\>{\it Un aspect de la question est politique}\\
\>\>?\>{\it Un c\^ot\a'e de la lettre est polisson}\\
\>\>\>{\it Une partie de l'\oe uvre est authentique}\\
\>\>?\>{\it Une phase de la fabrication est manuelle}\\
\>\>\>{\it Plusieurs traits de ce r\a'egime sont totalitaires}\end{tabbing}
The locative variant of {\it Vsup\/} =: {\it avoir\/} (\ref{25}) and the binding operator {\it avoir\/}
may account for sentences like:
\begin{tabbing}(000)\=(D) =\=?*\=\kill
\>\>\>{\it Det $N_0$ avoir ces\/} ({\it c\^ot\a'es\/} + {\it parties\/} + {\it traits\/}) (E + {\it en lu$i^0$\/})\\
\>\>\>{\it Det $N_0$ avoir ce\/} ({\it c\^ot\a'e\/} + {\it trait\/}) (E + {\it en lu$i^0$\/})\end{tabbing}

Fig~\ref{listeNapp}, at the end of the article, is a list of 267 nouns which provide examples
of the relations of Fig.~\ref{avoirNapp}. Each noun {\it Napp\/} is followed by one or more examples
of adjectives to which {\it Napp\/} is appropriate, and a few distributional properties. For example,
{\it acoustique\/} appears in the list because of sentences like
(\ref{11})--(\ref{16}). In each line, each distributional property is described by a sign, which is
a {\tt +} if the entry has the
property and a  {\tt -} if it does not. The properties are the following:
\begin{enumerate}
\item {\it $N_0$\/} =: {\it Nhum,} i.e. the subject of the {\it Vsup\/} construction may be a human noun.
\item {\it $N_0$\/} =: {\it N-hum,} the subject of the {\it Vsup\/} construction may be a non-human noun.
\item {\it Vsup\/} =: {\it avoir.}
\item {\it Vsup\/} =: {\it \^etre de.}
\item {\it Vsup\/} =: {\it r\'egner Loc.}
\item {\it Vsup\/} =: {\it faire.}
\item {\it Napp sing,} the appropriate noun may be singular if {\it $N_0$\/} is semantically singular.
\item {\it Napp plur,} the appropriate noun may be plural if {\it $N_0$\/} is semantically singular.
\item {\it Det $N_0$ Vsup de le Napp sing,} i.e. {\it Det Modif\/} commutes with {\it de le\/} with an intensive meaning.
\item {\it Det $N_0$ Vsup des Napp plur,} i.e. {\it Det Modif\/} commutes with {\it des\/} with an intensive meaning.
\item (A), i.e. (A) in Fig.~\ref{avoirNapp} is acceptable.
\item (D), i.e. (D) in Fig.~\ref{avoirNapp} is acceptable.
\item {\it Det $N_0$ Vsup quelque chose de Adj.}
\item {\it Det $N_0$ \^etre Adj Prep Det Napp.}
\end{enumerate}
The codes at the end of the lines identify syntactic tables of other authors where the same {\it Vsup\/} construction of the noun is described.

In a category of Korean nominal constructions, a noun {\it Napp,} e.g. {\it h\"angtong\/} "behaviour", {\it th\"ato\/} "attitude", {\it case\/} "appearance", appears as a "support" for an adjectival or adverbial predicate \cite[p.~196--211, 1994b]{Nam94a}. These constructions are characterized by the following properties:
\begin{itemize}
\item {\it Napp\/} is little or not autonomous:
\begin{tabbing}?*\=Minu Nomin \=[Ina towards \=Det \=behaviour] \=Acc \=do \=Past \=\kill
{\rm *}\>{\it Minu-n\^{\i}n}\>{\it [Ina-et\"aha}\>{\it -n}\>{\it h\"angtong]}\>{\it -\^{\i}l}\>{\it ha}\>{\it -\^os'}\>{\it -ta}\\
{\rm *}\>{\sl Minu Nomin}\>{\sl [Ina towards}\>{\sl Det}\>{\sl behaviour]}\>{\sl Acc}\>{\sl do}\>{\sl Past}\>{\sl Termin}\\
{\rm *}\>{\it Minu showed a behaviour towards Ina}\end{tabbing}
\begin{tabbing}?*\=Minu Nomin \=Ina-(-etahaso + -eke)- \=[behaviour] \=Acc \=do \=Past \=\kill
*\>{\it Minu-n\^{\i}n}\>{\it Ina\/}({\it -et\"ah\"as\^o\/} + {\it -eke\/})\>{\it [h\"angtong]}\>{\it -\^{\i}l}\>{\it ha}\>{\it -\^os'}\>{\it -ta}\\
{\rm *}\>{\sl Minu Nomin}\>{\sl Ina towards}\>{\sl [behaviour]}\>{\sl Acc}\>{\sl do}\>{\sl Past}\>{\sl Termin}\\
{\rm *}\>{\it Minu showed a behaviour towards Ina\/};\end{tabbing}
\item an adjectival or adverbial predicate is obligatory associated to {\it Napp\/}:
\begin{tabbing}?*\=Minu Nomin \=Ina-(-etahaso + -eke)- \=[very \=p'\^onp'\^ons\^{\i}l\^ow\=Det \=behaviour] \=\kill
\>{\it Minu-n\^{\i}n}\>{\it Ina\/}({\it -et\"ah\"as\^o\/} + {\it -eke\/})\>{\it [acu}\>{\it p'\^onp'\^ons\^{\i}l\^ow}\>{\it -n}\>{\it h\"angtong]}\>{\it -\^{\i}l}\\
\>{\sl Minu Nomin}\>{\sl Ina towards}\>{\sl [very}\>{\sl impudent}\>{\sl Det}\>{\sl behaviour]}\>{\sl Acc}\\
?*\=Minu\=do \=Past \=\kill
\>\>{\it ha}\>{\it -\^os'}\>{\it -ta}\\
\>\>{\sl do}\>{\sl Past}\>{\sl Termin}\\
\>{\it Minu showed a very impudent behaviour towards Ina}\end{tabbing}
\begin{tabbing}?*\=Minu Nomin \=Ina-(-etahaso + -eke)- \=very \=p'onp'onsilop\=Conj \=[behaviour] \=\kill
\>{\it Minu-n\^{\i}n}\>{\it Ina\/}({\it -et\"ah\"as\^o\/} + {\it -eke\/})\>{\it acu}\>{\it p'\^onp'\^ons\^{\i}l\^op}\>{\it -ke}\>{\it [h\"angtong]}\>{\it -\^{\i}l}\\
\>{\sl Minu Nomin}\>{\sl Ina towards}\>{\sl very}\>{\sl impudent}\>{\sl Conj}\>{\sl [behaviour]}\>{\sl Acc}\\
?*\=Minu\=do \=Past \=\kill
\>\>{\it ha}\>{\it -\^os'}\>{\it -ta}\\
\>\>{\sl do}\>{\sl Past}\>{\sl Termin}\\
\>{\it Minu showed a behaviour very impudently towards Ina\/};\end{tabbing}
\item when combined with an adverb, {\it Napp\/} cannot incorporate its complement into its noun phrase:
\begin{tabbing}?*\=Minu Nomin \=very \=p'onp'onsilop\=Conj \=[Ina towards \=Det \=behaviour] \=\kill
{\rm ?*}\>{\it Minu-n\^{\i}n}\>{\it acu}\>{\it p'\^onp'\^ons\^{\i}l\^op}\>{\it -ke}\>{\it [Ina-et\"aha}\>{\it -n}\>{\it h\"angtong]}\>{\it -\^{\i}l}\\
{\rm ?*}\>{\sl Minu Nomin}\>{\sl very}\>{\sl impudent}\>{\sl Conj}\>{\sl [Ina towards}\>{\sl Det}\>{\sl behaviour]}\>{\sl Acc}\\
?*\=Minu\=do \=Past \=\kill
\>\>{\it ha}\>{\it -\^os'}\>{\it -ta}\\
\>\>{\sl do}\>{\sl Past}\>{\sl Termin}\\
{\rm ?*}\>{\it Minu showed a behaviour towards Ina very impudently,}\end{tabbing}
whereas autonomous predicative nouns like {\it piphan\/} "criticism" can:
\begin{tabbing}?*\=Minu Nomin \=very \=kahokha\=Conj \=[Ina towards \=Det \=criticism] \=\kill
\>{\it Minu-n\^{\i}n}\>{\it acu}\>{\it kahokha}\>{\it -ke}\>{\it [Ina-et\"aha}\>{\it -n}\>{\it piphan]}\>{\it -\^{\i}l}\\
\>{\sl Minu Nomin}\>{\sl very}\>{\sl crual}\>{\sl Conj}\>{\sl [Ina towards}\>{\sl Det}\>{\sl criticism]}\>{\sl Acc}\\
?*\=Minu\=do \=Past \=\kill
\>\>{\it ha}\>{\it -\^os'}\>{\it -ta}\\
\>\>{\sl do}\>{\sl Past}\>{\sl Termin}\\
\>{\it Minu made a criticism towards Ina very crually\/};\end{tabbing}
\item {\it Napp\/} may be the nominalization of a {\it Vsup\/} like {\it h\"angtonghata\/} "behave":
\begin{tabbing}?*\=Minu Nomin \=Ina-(-etahaso + -eke)- \=very \=p'onp'onsilop\=\kill
\>{\it Minu-n\^{\i}n}\>{\it Ina\/}({\it -et\"ah\"as\^o\/} + {\it -eke\/})\>{\it acu}\>{\it p'\^onp'\^ons\^{\i}l\^op}\>{\it -ke}\\
\>{\sl Minu Nomin}\>{\sl Ina towards}\>{\sl very}\>{\sl impudent}\>{\sl Conj}\\
?*\=Minu\=h\"angtongha\=Past \=\kill
\>\>{\it h\"angtongha}\>{\it -\^os'}\>{\it -ta}\\
\>\>{\sl behave}\>{\sl Past}\>{\sl Termin}\\
\>{\it Minu behaved very impudently towards Ina,}\end{tabbing}
whereas autonomous predicative nouns may be the nominalization of distributional verbs like {\it piphanhata\/} "criticize":
\begin{tabbing}?*\=Minu Nomin \=Ina (towards + \=Acc) \=very \=kahokha\=Conj \=piphanha\=Past \=\kill
\>{\it Minu-n\^{\i}n}\>{\it Ina\/}({\it -et\"ah\"as\^o\/} +\>{\it -l\^{\i}l\/})\>{\it acu}\>{\it kahokha}\>{\it -ke}\>{\it piphanha}\>{\it -\^os'}\>{\it -ta}\\
\>{\sl Minu Nomin}\>{\sl Ina} ({\sl towards} +\>{\sl Acc})\>{\sl very}\>{\sl crual}\>{\sl Conj}\>{\sl criticize}\>{\sl Past}\>{\sl Term}\\
\>{\it Minu criticized Ina very crually}\end{tabbing}
\end{itemize}
These properties can be compared to those of the French nouns of this study, especially the obligatory
association with another predicate.
\subsection{Other constructions}

What the other entries of the corpus have in common is the unacceptability of:
\begin{tabbing}(000)\=(D) =\=?*\=\kill
\>(E)\>\>{\it Det Napp de Det $N_0$ \^etre Adj}\end{tabbing}
The analysis of Fig.~\ref{avoirNapp} is inadequate for them. Consider e.g. the noun {\it p\^ate\/} "stuff".
The left half of Fig.~\ref{avoirNapp} describes the following forms:
\begin{tabbing}(000)\=(D) =\=?*\=\kill
\>\>\>{\it Luc est d'une p\^ate qui est accommodante}\\
(\refstepcounter{numero}\arabic{numero}\label{33})
\>\>\>{\it Luc est d'une p\^ate accommodante}\end{tabbing}
There is no autonomous {\it Vsup\/} construction for {\it p\^ate\/}:
\begin{tabbing}(000)\=(D) =\=?*\=\kill
\>\>?\>{\it Luc est d'une certaine p\^ate}\end{tabbing}
If this sentence is acceptable, it has no relation with:
\begin{tabbing}(000)\=(D) =\=?*\=\kill
\>\>?\>{\it Marie conna\^{\i}t bien la p\^ate de Luc}\end{tabbing}
However, {\it Vsup\/} =: {\it \^etre de\/} acts as a {\it Vsup\/} for {\it Napp Adj\/} =: {\it p\^ate accommodante\/}:
\begin{tabbing}(000)\=(D) =\=?*\=\kill
\>\>?\>{\it Marie conna\^{\i}t bien la p\^ate accommodante de Luc}\end{tabbing}
Nevertheless, {\it Napp\/} =: {\it p\^ate\/} and {\it Adj\/} =: {\it accommodant\/} do not constitute a frozen idiomatic sequence,
since {\it accommodant\/} commutes with {\it d\'ebonnaire, serviable\/}\ldots

Now {\it Vsup Det Napp\/} =: {\it \^etre d'une p\^ate\/} commutes with {\it Vsup\/} =: {\it \^etre\/}:
\begin{tabbing}(000)\=(D) =\=?*\=\kill
(\refstepcounter{numero}\arabic{numero}\label{34})
\>\>\>{\it Luc est\/} ({\it accommodant\/} + {\it d\a'ebonnaire\/} + {\it serviable\/})\end{tabbing}
The appropriateness of {\it p\^ate\/} to (\ref{34}) appears in the relation (\ref{33})--(\ref{34}), i.e. (C)--(F), whereas in Fig.~\ref{avoirNapp} it was in the restructuration (E) = (F). The relation (\ref{33})--(\ref{34}) appears
as a substitution of supports. The sequence {\it Vsup Det Napp\/} =: {\it \^etre d'une p\^ate\/} plays the
part of a support for the adjective. We will use the term support noun, {\it Nsup,} in this section. The analysis of (\ref{33}) is shown in Fig.~\ref{avoirNsup1}.
\begin{figure}
\begin{tabbing}
(D) \=Det N0 avoir\=(D) \=un Npb qui \^etre Adj\=(D) \=\kill
(B)\>{\it Det $N_0$ Vsup Det Nsup qui \^etre Adj}\\
\>[{\it qui \^etre\/} z.]\\
(C)\>{\it Det $N_0$ Vsup Det Nsup Adj}\\
\>\>\>\>subst. supp.\>\\
\>\>\>\>(F)\>{\it Det $N_0$ \^etre Adj}
\end{tabbing}
\caption{Derivation of (27).}\label{avoirNsup1}
\end{figure}
In the construction (C) of Fig.~\ref{avoirNsup1}, both {\it Vsup\/} and {\it Vsup Det Nsup\/} have support status, as in the other examples in this section.

A more productive case is illustrated by {\it moment\/} in (\ref{35}) and by other nouns of time:
\begin{tabbing}(000)\=(D) =\=?*\=Luc\=\kill
(\refstepcounter{numero}\arabic{numero}\label{35})
\>\>\>{\it Luc conna\^{\i}t un moment\/} (E + {\it qui est\/}) {\it plein de\/} ({\it d\a'ecouragement}\\
\>\>\>\>+ {\it enthousiasme\/})\end{tabbing}
The construction (D) is rather acceptable:
\begin{tabbing}(000)\=(D) =\=?*\=Luc\=\kill
\>\>?\>({\it Certains\/} + {\it Les\/}) {\it moments que Luc conna\^{\i}t alors sont pleins de}\\
\>\>\>\>{\it d\a'ecouragement}
\end{tabbing}
whereas (E) is not:
\begin{tabbing}(000)\=(D) =\=?*\=\kill
\>\>*\>({\it Certains\/} + {\it Les\/}) {\it moments de Luc sont pleins de d\a'ecouragement}\end{tabbing}
Thus the construction (C) must be directly related to (F):
\begin{tabbing}(000)\=(D) =\=?*\=\kill
(\refstepcounter{numero}\arabic{numero}\label{38})
\>\>\>{\it Luc est plein de d\a'ecouragement\/} (E + {\it dans de tels moments\/}) \end{tabbing}
These relations are shown in Fig.~\ref{avoirNsup2}.
\begin{figure}
\begin{tabbing}
(D) \=Det N0 avoir\=(D) \=un Npb qui \^etre Adj\=(D) \=\kill
(B)\>{\it Det $N_0$ Vsup un Nsup qui \^etre Adj}\\
\>\>\>\>(D)\>{\it Det Nsup que $N_0$ Vsup \^etre Adj}\\
\>[{\it qui \^etre\/} z.]\\
(C)\>{\it Det $N_0$ Vsup un Nsup Adj}\\
\>\>\>\>subst. supp.\>\\
\>\>\>\>(F)\>{\it Det $N_0$ \^etre Adj\/} (E + {\it Prep Det Nsup\/})
\end{tabbing}
\caption{Derivation of (30).}\label{avoirNsup2}
\end{figure}

The last situation that we examine is that of e.g. {\it caract\`ere\/} "feature" in:
\begin{tabbing}(000)\=(D) =\=?*\=\kill
(\refstepcounter{numero}\arabic{numero}\label{39})
\>\>\>{\it Ces affirmations ont un caract\a`ere diffamatoire}\\
(\refstepcounter{numero}\arabic{numero}\label{40})
\>=\>\>{\it Ces affirmations sont diffamatoires}\end{tabbing}
Elementary sentences of the following types can be observed:
\begin{tabbing}(000)\=(D) =\=?*\=\kill
\>\>?\>{\it Ces affirmations ont certains caract\a`eres}\\
\>\>?\>{\it Ces affirmations ont ce caract\a`ere}\end{tabbing}
They are not entirely autonomous, like (\ref{10}). Their {\it Napp\/} cannot become the subject of
{\it \^etre Adj.} Thus, none of the constructions (A), (B), (D), (E) is acceptable:
\begin{tabbing}(000)\=(D) =\=?*\=Ces\=\kill
\>(A)\>*\>{\it Ces affirmations ont un certain caract\a`ere, ce caract\a`ere est}\\
\>\>\>\>{\it diffamatoire}\\
\>(B)\>*\>{\it Ces affirmations ont un caract\a`ere qui est diffamatoire}\\
\>(B)\>*\>{\it Un certain caract\a`ere qu'ont ces affirmations est diffamatoire}\\
\>(E)\>*\>{\it Un certain caract\a`ere de ces affirmations est diffamatoire}\end{tabbing}
These nouns are clearly supports of the adjective (Fig.~\ref{avoirNsup3}).
\begin{figure}
\begin{tabbing}
(D) \=Det N0 avoir\=(D) \=un Npb qui \^etre Adj\=(D) \=\kill
(C)\>{\it Det $N_0$ Vsup Det Nsup Adj}\\
\>\>\>\>subst. supp.\>\\
\>\>\>\>(F)\>{\it Det $N_0$ \^etre Adj\/} (E + {\it de Nsup\/})
\end{tabbing}
\caption{Derivation of (32) and similar sentences.}\label{avoirNsup3}
\end{figure}
Fig~\ref{listeNsup} is a list of 18 nouns which provide examples of the relations of Fig.~\ref{avoirNsup1}, \ref{avoirNsup2} and \ref{avoirNsup3}. Each noun {\it Nsup\/} is followed by one or more examples of adjectives to which {\it Nsup\/} is appropriate, and a few distributional properties. For example, {\it caract\`ere\/} appears in the list because of sentences like
(\ref{39})--(\ref{40}). In each line, each distributional property is described by a sign, which is a {\tt +} if the entry has the
property and a  {\tt -} if it does not. The properties are the following:
\begin{enumerate}
\item {\it $N_0$\/} =: {\it Nhum,} i.e. the subject of the {\it Vsup\/} construction may be a human noun.
\item {\it $N_0$\/} =: {\it N-hum,} the subject of the {\it Vsup\/} construction may be a non-human noun.
\item {\it Vsup\/} =: {\it avoir.}
\item {\it Vsup\/} =: {\it \^etre de.}
\item {\it Vsup\/} =: {\it faire.}
\item {\it Nsup sing,} the appropriate noun may be singular if {\it $N_0$\/} is semantically singular.
\item {\it Nsup plur,} the appropriate noun may be plural if {\it $N_0$\/} is semantically singular.
\item (B), i.e. (B) is acceptable.
\item (D), i.e. (D) is acceptable.
\item {\it Det $N_0$ Vsup quelque chose de Adj.}
\item {\it Det $N_0$ \^etre Adj Prep Det Nsup.}
\end{enumerate}
The codes at the end of the lines identify syntactic tables of other authors where the same {\it Vsup\/} construction of the noun is described.
\begin{figure}
\begin{tabbing}
citoyennete de \=maniaque, fievreux, paludeen de \= accommodant \=\kill
{\it Nsup}\>{\it Adj}\>{\tt 12345678901}\>{\sf Tables}\\
\\
{\it acc\a`es}\>{\it maniaque, fi\a'evreux, palud\a'een}\>{\tt +-+--++---+}\\
{\it caract\a`ere}\>{\it diffamatoire}\>{\tt +++--++--+-}\\
{\it citoyennet\a'e}\>{\it danois}\>{\tt +-+--++---+}\>{\sf APP1 DR3 AN07}\\
{\it confession}\>{\it protestant}\>{\tt +-++-+-----}\>{\sf AN07}\\
{\it dimension}\>{\it politique}\>{\tt -++--+---++}\>{\sf APP1 AN08}\\
{\it expression}\>{\it francophone}\>{\tt ++++-+-----}\>{\sf AN07}\\
{\it go\^ut}\>{\it douteux, infect}\>{\tt -+++-+-+-+-}\>{\sf AN07}\\
{\it heure}\>{\it plein d'angoisse}\>{\tt +-+--++++--}\\
{\it heure}\>{\it angoiss\a'e}\>{\tt +-+---+++--}\\
{\it jour}\>{\it plein de bonheur}\>{\tt +-+---+++--}\>{\sf AN10}\\
{\it jour}\>{\it heureux}\>{\tt +-+---+----}\>{\sf AN10}\\
{\it moment}\>{\it charmant, heureux}\>{\tt +-+--++++-+}\>{\sf AN10}\\
{\it naissance}\>{\it cr\a'eole, pied-noir}\>{\tt +-++-+----+}\\
{\it origine}\>{\it indig\a`ene}\>{\tt +-++-+++--+}\\
{\it parole}\>{\it optimiste}\>{\tt +-+--+-++--}\\
{\it p\^ate}\>{\it accommodant}\>{\tt +-++-+-+-+-}\>{\sf AN07}\\
{\it p\a'eriode}\>{\it mystique}\>{\tt +-+--++--++}\>{\sf AN10}\\
{\it phase}\>{\it mystique}\>{\tt +++--++--++}
\end{tabbing}
\caption{List of {\it Nsup}'s.}\label{listeNsup}
\end{figure}

\subsection{Unrelated constructions}

This study is {\em not\/} concerned with sentences which are superficially similar to (C),
but whose verb is not a support verb. The following sentence:
\begin{tabbing}(000)\=(D) =\=?*\=\kill
(\refstepcounter{numero}\arabic{numero}\label{41})
\>\>\>{\it Les vols ont l'air fr\a'equents dans ce magasin}\end{tabbing}
is not a {\it Vsup\/} construction:
\begin{tabbing}(000)\=(D) =\=?*\=\kill
\>\>*\>{\it L'air des vols est fr\a'equent dans ce magasin}\\
\>\>*\>{\it On constate l'air fr\a'equent des vols dans ce magasin}\end{tabbing}
(\ref{41}) is related to:
\begin{tabbing}(000)\=(D) =\=?*\=\kill
\>\>\>{\it Les vols ont l'air d'\^etre fr\a'equents dans ce magasin}\end{tabbing}
where {\it avoir l'air de\/} "to seem to" is a compound verb described in the table C8 of M.~Gross.

\section{Prepositional complements of {\it Napp\/} and {\it Adj}}

In the preceding sections we did not consider the prepositional complements of the nouns and adjectives.
We examine here the behaviour of these complements in the various forms of the sentences. 

\subsection{Essential complements of {\it Adj}} 

Let us call adjectival sentences those whose predicate is {\it \^etre Adj.} Their general form is:
\begin{tabbing}(000)\=(D) =\=?*\=\kill
\>\>\>{\it Det $N$ \^etre Adj W}\end{tabbing}
where {\it W\/} stands for a (possibly empty) sequence of essential complements. At least two of the constructions
in Fig.~\ref{avoirNpc} and \ref{avoirNapp} are adjectival sentences:
the second sentence of (A), namely:
\begin{tabbing}(000)\=(D) =\=?*\=\kill
\>\>\>{\it Det $Napp$ \^etre Adj W,}\end{tabbing}
and (F):
\begin{tabbing}(000)\=(D) =\=?*\=\kill
\>\>\>{\it Det $N_0$ \^etre Adj $W^\prime$}\end{tabbing}
A systematic comparison of $W$ and $W^\prime$ shows that whenever the adjectival sentence in
(A) contains one or more essential complements, these complements are preserved along
the whole derivation from (A) to (F). Thus we have with an {\it Npb\/}:
\begin{tabbing}(000)\=(D) =\=?*\=\kill
\>\>?*\>{\it Luc a une peau, cette peau est moite de sueur}\\
\>\>?*\>{\it La peau qu'a Luc est moite de sueur}\\
\>\>\>{\it La peau de Luc est moite de sueur}\\
\>\>\>{\it Luc est moite de sueur}\end{tabbing}
and with an {\it Napp\/} from Fig.~\ref{listeNapp}:
\begin{tabbing}(000)\=(D) =\=?*\=Sa \=\kill
\>\>\>{\it Sa venue a une certaine date, cette date est ant\a'erieure \a`a ceci}\\
\>\>\>\>{\it d'un mois}\\
\>\>\>{\it La date qu'a sa venue est ant\a'erieure \a`a ceci d'un mois}\\
\>\>\>{\it La date de sa venue est ant\a'erieure \a`a ceci d'un mois}\\
\>\>\>{\it Sa venue est ant\a'erieure \a`a ceci d'un mois}\end{tabbing}
At first sight, there exist cases where no essential complement can occur  in
(A), (D), (E), but where an essential complement is observed in (F):
\begin{tabbing}(000)\=(D) =\=?*\=Luc a un certain temperament, \=\kill
(\refstepcounter{numero}\arabic{numero}\label{42})
\>(A)\>\>{\it Luc a un certain temp\a'erament, ce temp\a'erament est jaloux}\\
\>\>\>\>(E + *{\it des succ\a`es de Paul\/})\\
(\refstepcounter{numero}\arabic{numero}\label{43})
\>(D)\>\>{\it Le temp\a'erament qu'a Luc est jaloux\/} (E + *{\it des succ\a`es de Paul\/})\\
(\refstepcounter{numero}\arabic{numero}\label{44})
\>(E)\>\>{\it Le temp\a'erament de Luc est jaloux\/} (E + *{\it des succ\a`es de Paul\/})\\
\>(F)\>\>{\it Luc est jaloux\/} (E + {\it des succ\a`es de Paul\/}),\end{tabbing}
In other words, {\it temp\'erament\/} is appropriate to:
\begin{tabbing}(000)\=(D) =\=?*\=\kill
(\refstepcounter{numero}\arabic{numero}\label{45})
\>\>\>{\it Luc est jaloux}\end{tabbing}
but not to:
\begin{tabbing}(000)\=(D) =\=?*\=\kill
(\refstepcounter{numero}\arabic{numero}\label{46})
\>\>\>{\it Luc est jaloux des succ\a`es de Paul}\end{tabbing}
This fact suggests a distinction between two entries of the adjective, an entry
\begin{tabbing}(000)\=(D) =\=?*\=\kill
\>\>\>{\it Det Napp \^etre jaloux}\end{tabbing}
which accounts for (\ref{42}), (\ref{43}), (\ref{44}), (\ref{45}), and an entry
\begin{tabbing}(000)\=(D) =\=?*\=\kill
\>\>\>{\it Det Nhum \^etre jaloux de\/} ({\it ce que P\/} + {\it Det N\/})\end{tabbing}
which enters only in the constructions of (\ref{45}) and (\ref{46}).
This distinction is more or less
satisfactory, depending on the adjective. In the case of {\it jaloux\/} "jealous", {\it heureux\/} "happy" and many other
adjectives of feelings, the distinction is correlated with a slight aspectual difference,
and the sentence (\ref{45}) is perhaps ambiguous.
For other adjectives, the distinction separates two entries with
rather different properties. Consider e.g. {\it rapide\/} "quick" in the following sentences:
\begin{tabbing}(000)\=(D) =\=?*\=\kill
(\refstepcounter{numero}\arabic{numero}\label{47})
\>(A)\>\>{\it Le convoi a une certaine progression, cette progression est rapide}\\
(\refstepcounter{numero}\arabic{numero}\label{48})
\>(D)\>\>{\it La progression qu'a le convoi est rapide}\\
(\refstepcounter{numero}\arabic{numero}\label{49})
\>(E)\>\>{\it La progression du convoi est rapide}\\
(\refstepcounter{numero}\arabic{numero}\label{50})
\>(F)\>\>{\it Le convoi est rapide}\end{tabbing}
No essential complements can be inserted into (\ref{47}), (\ref{48}), (\ref{49}). A complement
can be inserted into (\ref{50}):
\begin{tabbing}(000)\=(D) =\=?*\=\kill
\>\>\>{\it Le convoi est rapide \a`a s'arr\^eter}\end{tabbing}
but, in that case, an appropriate noun is no longer available, and the distribution of $N_0$
is much wider than in (\ref{50}):
\begin{tabbing}(000)\=(D) =\=?*\=\kill
\>\>\>{\it L'eau a \a'et\a'e rapide \a`a s'\a'evaporer}\\
\>\>\>{\it Le danger a \a'et\a'e rapide \a`a augmenter}\end{tabbing}
The expression {\it \^etre rapide \`a\/} can even be inserted into a frozen sentence with a frozen subject:
\begin{tabbing}(000)\=(D) =\=?*\=\kill
\>\>\>{\it La moutarde a \a'et\a'e rapide \a`a monter au nez de Luc}\end{tabbing}
Thus, the entry
\begin{tabbing}(000)\=(D) =\=?*\=\kill
\>\>\>{\it Det Napp \^etre rapide}\end{tabbing}
is distinguished from the entry
\begin{tabbing}(000)\=(D) =\=?*\=\kill
\>\>\>{\it Det $N_0$ \^etre rapide \a`a $V^0$inf W}\end{tabbing}
on syntactic grounds, and this syntactic distinction corroborates our semantic intuitions.

A similar situation occurs with verbs like {\it tourmenter\/} "torment" \cite[table 4]{Gross75}.
The nominal distributions of the subject and of the direct object are preserved by the passive
transformation:
\begin{tabbing}(000)\=(D) =\=?*\=\kill
\>\>\>({\it Luc\/} + {\it Ce sentiment\/} + {\it Ceci\/}) {\it tourmente L\a'ea}\\
\>\>\>{\it L\a'ea est tourment\a'ee\/} (E + {\it par\/} ({\it Luc\/} + {\it ce sentiment\/} + {\it ceci\/}))\end{tabbing}
The passive without agent occurs with an appropriate subject {\it caract\`ere\/} "character":
\begin{tabbing}(000)\=(D) =\=?*\=\kill
(\refstepcounter{numero}\arabic{numero}\label{51})
\>(E)\>\>{\it Le caract\a`ere de L\a'ea est tourment\a'e}\\
(\refstepcounter{numero}\arabic{numero}\label{52})
\>(F) =\>\>{\it L\a'ea est tourment\a'ee}\end{tabbing}
However, {\it caract\`ere\/} is forbidden in the direct object of the active and in the
subject of the passive with agent:
\begin{tabbing}(000)\=(D) =\=?*\=\kill
\>\>*\>({\it Luc\/} + {\it Ce sentiment\/} + {\it Ceci\/}) {\it tourmente le caract\a`ere de L\a'ea}\\
\>\>*\>{\it Le caract\a`ere de L\a'ea est tourment\a'e par\/} ({\it Luc\/} + {\it ce sentiment\/} + {\it ceci\/})\end{tabbing}
An adjectival entry {\it tourment\'e,} with no essential complements, can account for (\ref{51}) and (\ref{52}).
This adjectival entry may be without synchronic connection with {\it tourmenter,} i.e. (\ref{51}) and (\ref{52}) may not be passive constructions.

If we consider specifically the entries of {\it jaloux, heureux, rapide, tourment\'e\/} which take all the forms 
(D), (E), (F), we observe that none of these forms has any essential complement.
In fact, we did not encounter any example where an essential complement
would be observed in (F) but not  in (A), (D), (E). If there is not,
the essential complements of the adjective are preserved in the whole Fig.~\ref{avoirNapp},
and conversely any essential complement in (F) has its source in (A).
If we include an essential complement {\it W\/} of the adjective in Fig.~\ref{avoirNapp}, we obtain Fig.~\ref{avoirNappAdj}.
\begin{figure}
\begin{tabbing}
(D) \=N0 avoir\=(D) \=un Npb qui \^etre Adj (D)\=(D) \=\kill
\>\>(A)\>{\it $N_0$ Vsup un certain Napp, ce Napp \^etre Adj W}\\
\>\>Rel.\>\>Rel.\\
(B)\>{\it $N_0$ Vsup un Napp qui \^etre Adj W}\>\>\>(D)\>{\it Napp que $N_0$ Vsup \^etre Adj W}\\
\>[{\it qui \^etre\/} z.]\>\>\>\>[Red. {\it Vsup\/}]\\
(C)\>{\it $N_0$ Vsup un Napp Adj W}\>\>\>(E)\>{\it Napp de $N_0$ \^etre Adj W}\\
\>\>\>\>\>[Restruct.]\\
\>\>\>\>(F)\>{\it $N_0$ \^etre Adj W\/} (E + {\it Prep Napp\/})\end{tabbing}
\caption{Essential complements of {\it Adj.} Free determiners are omitted.}\label{avoirNappAdj}
\end{figure}

\subsection{Essential complements of {\it Napp}} 

The first sentence of (A) in Fig.~\ref{avoirNapp} is a {\it Vsup\/} construction with a nominal predicate.
Such a construction can contain a prepositional complement. If it does, the predicative noun and the
complement often have the property of "double analysis", i.e. they can be considered both as one constituent of the sentence
and as a sequence of two constituents \cite{Giry78b}, since they can be extracted either jointly or separately.
In this section and in the next one we examine whether the complement appears
in the other forms of Fig.~\ref{avoirNapp}, namely (B), (C), (D), (E) and (F). We will consider separately 
essential complements and adverbial complements, though this distinction is often not evident \cite[1984]{Labelle83}.

A few nouns from Fig.~\ref{listeNapp} clearly have an essential complement. The preposition, the distribution of the complement, and its interpretation are specific to the {\it Napp\/}:
\begin{tabbing}(000)\=(D) =\=?*\=\kill
\>\>\>{\it Cette colle a une certaine adh\a'erence au m\a'etal}\\
\>\>\>{\it Cet os a une certaine datation par les sp\a'ecialistes}\\
\>\>\>{\it Luc a une certaine d\a'etermination \a`a partir}\\
\>\>\>{\it Luc a une certaine fa\c{c}on de parler}\\
\>\>\>{\it Luc a une certaine mani\a`ere de parler}\\
\>\>\>{\it Le convoi a une certaine progression vers la mer}\\
\>\>\>{\it Luc tient certains propos \a`a L\a'ea}\\
\>\>\>{\it Luc a une certaine volont\a'e de partir}\end{tabbing}
The complement does not have the mobility of an adverb:
\begin{tabbing}(000)\=(D) =\=?*\=\kill
\>\>*\>{\it \a`A partir, Luc a une certaine d\a'etermination}\\
\>\>*\>{\it Luc, \a`a partir, a une certaine d\a'etermination}\\
\>\>?\>{\it Luc a \a`a partir une certaine d\a'etermination}\end{tabbing}
For these nouns, the essential complement is preserved in all the forms of Fig.~\ref{avoirNapp} whenever
{\it Napp\/} appears, i.e. except in the shorter form of (F):
\begin{tabbing}(000)\=(D) =\=?*\=Luc\=\kill
\>(A)\>\>{\it Luc a une certaine d\a'etermination \a`a partir, cette d\a'etermination}\\
\>\>\>\>{\it est inflexible}\\
\>(B)\>\>{\it Luc a une d\a'etermination \a`a partir qui est inflexible}\\
\>(C)\>\>{\it Luc a une d\a'etermination \a`a partir inflexible}\\
\>(D)\>\>{\it La d\a'etermination \a`a partir qu'a Luc est inflexible}\\
\>(E)\>\>{\it La d\a'etermination \a`a partir de Luc est inflexible}\\
\>(F$_1$)\>\>{\it Luc est inflexible dans sa d\a'etermination \a`a partir}\\
\>(F$_2$)\>*\>{\it Luc est inflexible \a`a partir}\end{tabbing}
The place of the complement in the examples above reflects the analysis where {\it Napp\/} and its complement
make up a constituent of the sentence and are extracted jointly when the relative of (D) is formed.
In (D) and (E), the complement can also take a place which is consistent with the dislocated analysis
where {\it Napp\/} is extracted separately:
\begin{tabbing}(000)\=(D) =\=?*\=\kill
\>(D)\>\>{\it La d\a'etermination qu'a Luc \a`a partir est inflexible}\\
\>(E)\>\>{\it La d\a'etermination de Luc \a`a partir est inflexible}\end{tabbing}
This dislocation is also observed in (C) and perhaps in (B):
\begin{tabbing}(000)\=(D) =\=?*\=\kill
\>(B)\>?\>{\it Luc a une d\a'etermination qui est inflexible \a`a partir}\\
\>(C)\>\>{\it Luc a une d\a'etermination inflexible \a`a partir}\end{tabbing}
In the case of {\it propos\/} "words" which has only the dislocated analysis, only the dislocated forms seem
acceptable:
\begin{tabbing}(000)\=(D) =\=?*\=\kill
\>(A)\>\>{\it Luc a tenu certains propos \a`a L\a'ea, ces propos sont grandiloquents}\\
\>(B)\>\>{\it Luc a tenu des propos qui sont grandiloquents \a`a L\a'ea}\\
\>(C)\>\>{\it Luc a tenu des propos grandiloquents \a`a L\a'ea}\\
\>(D)\>\>{\it Les propos que Luc a tenus \a`a L\a'ea sont grandiloquents}\\
\>(E)\>\>{\it Les propos de Luc \a`a L\a'ea sont grandiloquents}\\
\>(F$_1$)\>\>{\it Luc a \a'et\a'e grandiloquent dans ses propos \a`a L\a'ea}\\
\>(F$_2$)\>*\>{\it Luc a \a'et\a'e grandiloquent \a`a L\a'ea}\end{tabbing}
If we include an essential complement {\it W\/} of {\it Napp\/} in Fig.~\ref{avoirNapp}, we obtain Fig.~\ref{avoirNappEss} in the case of a nominal construction with double analysis.
\begin{figure}
\begin{tabbing}
(D) \=N0 avoir\=(D) \=un Npb qui \^etre Adj (D)\=(D) \=\kill
\>\>(A)\>{\it $N_0$ Vsup un certain Napp  W, ce Napp \^etre Adj}\\
\>\>Rel.\>\>Rel.\\
(B)\>{\it $N_0$ Vsup un Napp W qui \^etre Adj}\>\>\>(D)\>{\it Napp W que $N_0$ Vsup \^etre Adj}\\
(B)\>{\it $N_0$ Vsup un Napp qui \^etre Adj W}\>\>\>(D)\>{\it Napp que $N_0$ Vsup W \^etre Adj}\\
\>[{\it qui \^etre\/} z.]\>\>\>\>[Red. {\it Vsup\/}]\\
(C)\>{\it $N_0$ Vsup un Napp W Adj}\>\>\>(E)\>{\it Napp W de $N_0$ \^etre Adj}\\
(C)\>{\it $N_0$ Vsup un Napp Adj W}\>\>\>(E)\>{\it Napp de $N_0$ W \^etre Adj}\\
\>\>\>\>\>[Restruct.]\\
\>\>\>\>(F)\>{\it $N_0$ \^etre Adj\/} (E + {\it Prep Napp W\/})\end{tabbing}
\caption{Essential complements of {\it Napp.} Free determiners are omitted.}\label{avoirNappEss}
\end{figure}

\subsection{Adverbial complements of {\it Napp}} 

Several other families of prepositional complements can occur in the first sentence of (A).
They have adverbial properties and have been mentioned as supporting the relation (E)--(F)
\cite{Vives82}. One of these families of complements takes the following forms:
\begin{tabbing}(000)\=(D) =\=?*\=\kill
\>\>\>({\it envers\/} + {\it \a`a l'\a'egard de\/} + {\it \a`a l'endroit de\/}
+ {\it vis-\a`a-vis de\/}) {\it Det Nhum}\end{tabbing}
These complements have adverbial properties:
\begin{itemize}
\item the preposition is not fixed;
\item they have the mobility of adverbs:
\begin{tabbing}(000)\=?*\=\kill
\>\>{\it Envers L\a'ea, Luc a une certaine attitude}\\
\>\>{\it Luc, envers L\a'ea, a une certaine attitude}\\
\>\>{\it Luc a envers L\a'ea une certaine attitude}\\
\>\>{\it Luc a une certaine attitude envers L\a'ea\/};\end{tabbing}
\item they are not specific to a given predicate, but combine with a more or less wide set
of predicates:
\begin{tabbing}(000)\=?*\=Luc a un certain\=\kill
\>\>{\it Luc a un certain\/} ({\it comportement\/} + {\it empressement\/} + {\it langage}\\
\>\>\> + {\it morgue\/} + {\it sentiment\/} + {\it z\a`ele\/}) {\it envers L\a'ea}\end{tabbing}
and they still have the same interpretation with these predicates. This selection or condition of
compatibility takes place
between {\it Napp\/} and the prepositional complement, and not between {\it Adj\/} and the complement
\cite{Vives82}\footnote{If {\it Adj\/} is replaced by another {\it Adj\/} for a given {\it Napp,} the
selection is not modified, but if {\it Napp\/} is replaced by another {\it Napp\/} for a given {\it Adj,}
the selection is modified.};
\item they combine with other complements with the same adverbial properties:
\begin{tabbing}(000)\=?*\=\kill
\>\>{\it Luc a une certaine attitude envers L\a'ea sur ce point aux yeux de Jo\/};\end{tabbing}
\item they are not obligatory:
\begin{tabbing}(000)\=?*\=\kill
\>\>{\it Luc a une certaine attitude\/} (E + {\it envers L\a'ea\/})\end{tabbing}
\end{itemize}
There are other families of complements with the same properties:
\begin{itemize}
\item ({\it avec\/} + {\it dans Pos$s^0$ relations avec\/}) {\it Det Nhum}
\item ({\it aux yeux de\/} + {\it pour\/}) {\it Det Nhum}
\item ({\it devant\/} + {\it face \a`a\/}) {\it Det Nhum}
\item ({\it sur\/} + {\it \a`a l'\a'egard de\/} + {\it vis-\a`a-vis de\/} + {\it quant \a`a\/}) ({\it ce point\/} + {\it Det Nnr\/})
\item scenic locative complements, e.g. in:
\begin{tabbing}(000)\=?*\=\kill
\>\>{\it Le fleuve a un certain trac\a'e dans la for\^et}\end{tabbing}
\end{itemize}
All of these adverbial complements provide examples of "double analysis":
\begin{tabbing}(000)\=(D) =\=?*\=\kill
(\refstepcounter{numero}\arabic{numero}\label{53})
\>\>\>{\it Voici l'attitude envers L\a'ea qu'a Luc}\\
(\refstepcounter{numero}\arabic{numero}\label{54})
\>\>\>{\it Voici l'attitude qu'a Luc envers L\a'ea}\end{tabbing}
In the following, the forms which reflect the dislocated analysis, e.g. (\ref{54}), are preferred to those which reflect the other analysis (\ref{53}).

The adverbial complements are preserved in all the forms of Fig.~\ref{avoirNapp}, 
including in the shorter form of (F):
\begin{tabbing}(000)\=(D) =\=?*\=\kill
(\refstepcounter{numero}\arabic{numero}\label{57})
\>(A)\>\>{\it Luc a une certaine attitude avec L\a'ea, cette attitude est ferme}\\
\>(B)\>\>{\it Luc a une attitude avec L\a'ea qui est ferme}\\
\>(C)\>\>{\it Luc a une attitude avec L\a'ea ferme}\\
\>(D)\>\>{\it L'attitude que Luc a avec L\a'ea est ferme}\\
\>(E)\>\>{\it L'attitude de Luc avec L\a'ea est ferme}\\
\>(F$_1$)\>\>{\it Luc est ferme dans son attitude avec L\a'ea}\\
(\refstepcounter{numero}\arabic{numero}\label{56})
\>(F$_2$)\>\>{\it Luc est ferme avec L\a'ea}\end{tabbing}
The derivation (A)=(D)=(E)=(F$_1$)=(F$_2$) is due to \cite{Vives82}. In (B) and (C),
the complement can move after the adjective:
\begin{tabbing}(000)\=(D) =\=?*\=\kill
\>(B)\>\>{\it Luc a une attitude qui est ferme avec L\a'ea}\\
\>(C)\>\>{\it Luc a une attitude ferme avec L\a'ea}\end{tabbing}
The locative complement of {\it rang\/} "rank" must probably be considered as an essential complement since its intepretation is
specific to this noun. However, unlike the essential complements of the preceding section, it has the mobility of an adverb and is preserved in the shorter form of (F):
\begin{tabbing}(000)\=(D) =\=?*\=\kill
\>(A)\>\>{\it Luc a un certain rang dans cette hi\a'erarchie, ce rang est subalterne}\\
\>(E)\>\>{\it Le rang de Luc dans cette hi\a'erarchie est subalterne}\\
\>(F$_2$)\>\>{\it Luc est subalterne dans cette hi\a'erarchie}\end{tabbing}

\subsection{Adverbial complements of {\it Adj}} 

Adverbial complements can also occur in the adjectival sentence of (A), they are preserved
in all forms:
\begin{tabbing}(000)\=(D) =\=?*\=\kill
(\refstepcounter{numero}\arabic{numero}\label{58})
\>(A)\>\>{\it Luc a une certaine allure, cette allure est louche aux yeux de L\a'ea}\\
\>(B)\>\>{\it Luc a une allure qui est louche aux yeux de L\a'ea}\\
\>(C)\>\>{\it Luc a une allure louche aux yeux de L\a'ea}\\
\>(D)\>\>{\it L'allure qu'a Luc est louche aux yeux de L\a'ea}\\
\>(E)\>\>{\it L'allure de Luc est louche aux yeux de L\a'ea}\\
\>(F$_1$)\>\>{\it Luc est louche aux yeux de L\a'ea par son allure}\\
(\refstepcounter{numero}\arabic{numero}\label{59})
\>(F$_2$)\>\>{\it Luc est louche aux yeux de L\a'ea}\end{tabbing}
Of course, a selection is observed between {\it Adj\/} and the prepositional complement.
Surprisingly, this complement can move out of the modifier in (B) and (C):
\begin{tabbing}(000)\=(D) =\=?*\=\kill
\>(B)\>\>{\it Luc, aux yeux de L\a'ea, a une allure qui est louche}\\
\>(C)\>\>{\it Luc, aux yeux de L\a'ea, a une allure louche}\end{tabbing}

Consequently, when an adverbial complement occurs in the shorter form of (F),
it may originate either from the first sentence of (A), e.g. (\ref{57})--(\ref{56}), 
or from the adjectival sentence of (A), e.g. (\ref{58})--(\ref{59}). The place of
the complement reflects this origin in (B) and (C):
\begin{tabbing}(000)\=(D) =\=?*\=\kill
\>(B)\>\>{\it Luc a une attitude avec L\a'ea qui est ferme}\\
\>(B)\>?*\>{\it Luc a une allure aux yeux de L\a'ea qui est louche}\end{tabbing}
and in (D) and (E):
\begin{tabbing}(000)\=(D) =\=?*\=\kill
\>(D)\>?*\>{\it L'attitude qu'a Luc est ferme avec L\a'ea} \cite{Vives82}\\
\>(D)\>\>{\it L'allure qu'a Luc est louche aux yeux de L\a'ea}\end{tabbing}
It is often the case that one and the same adverbial complement can occur in
both sentences of (A). Then, surprisingly, the meaning of the sentences is
independent of the origin of the complement:
\begin{tabbing}(000)\=(D) =\=?*\=\kill
\>(A)\>\>{\it Luc a une certaine attitude envers L\a'ea, cette attitude est courtoise}\\
\>(E)\>\>{\it L'attitude de Luc envers L\a'ea est courtoise}\\
\>(A)\>\>{\it Luc a une certaine attitude, cette attitude est courtoise envers L\a'ea}\\
\>(E)\>\>{\it L'attitude de Luc est courtoise envers L\a'ea}\end{tabbing}
and though this analysis assigns two sources to (F$_2$) simultaneously, the latter is not perceptibly
ambiguous:
\begin{tabbing}(000)\=(D) =\=?*\=\kill
\>(F$_2$)\>\>{\it Luc est courtois envers L\a'ea}\end{tabbing}
To solve this difficulty, we suggest considering that the canonical place of the complement
is in the first sentence of (A), and that it can move into the adjectival sentence
if a selection between the complement and the adjective allows it. If we include
an adverbial complement $W_1$ of {\it Napp\/} and an adverbial complement $W_2$ of {\it Adj\/} in
Fig.~\ref{avoirNapp}, we obtain Fig.~\ref{avoirNappAdv}.
\begin{figure}[tb]
\begin{tabbing}
(D) \=N0 avoir\=(D) \=un Npb qui \^etre Adj\=(D) \=(D)\=(D) \=\kill
\>\>(A)\>{\it $N_0$ Vsup un certain Napp $W_1$, ce Napp \^etre Adj $W_2$}\\
\>\>Rel.\>\>Rel.\\
\>\>\>\>(D)\>{\it Napp que $N_0$ Vsup $W_1$ \^etre Adj $W_2$}\\
\>\>\>\>(D)\>{\it Napp $W_1$ que $N_0$ Vsup \^etre Adj $W_2$}\\
\>\>\>\>(D)\>{\it Napp que $N_0$ Vsup $W_1$ $W_2$ \^etre Adj}\\
\>\>\>\>(D)\>{\it Napp $W_1$ que $N_0$ Vsup $W_2$ \^etre Adj}\\
\>\>\>\>\>\>[Red. {\it Vsup\/}]\\
(B)\>{\it $N_0$ Vsup un Napp $W_1$ qui \^etre Adj $W_2$}\>\>\>\>\>(E)\>{\it Napp de $N_0$ $W_1$ \^etre Adj $W_2$}\\
(B)\>{\it $N_0$ Vsup un Napp qui \^etre Adj $W_2$ $W_1$}\>\>\>\>\>(E)\>{\it Napp $W_1$ de $N_0$ \^etre Adj $W_2$}\\
(B)\>{\it $N_0$ Vsup $W_2$ un Napp $W_1$ qui \^etre Adj}\>\>\>\>\>(E)\>{\it Napp de $N_0$ $W_1$ $W_2$ \^etre Adj}\\
(B)\>{\it $N_0$ Vsup $W_2$ un Napp qui \^etre Adj $W_1$}\>\>\>\>\>(E)\>{\it Napp $W_1$ de $N_0$ $W_2$ \^etre Adj}\\
\>[{\it qui \^etre\/} z.]\>\>\>\>\>\>[Restruct.]\\
(C)\>{\it $N_0$ Vsup un Napp $W_1$ Adj $W_2$}\>\>\>\>\>(F$_1$)\>{\it $N_0$ \^etre Adj $W_2$ Prep Napp $W_1$}\\
(C)\>{\it $N_0$ Vsup un Napp Adj $W_2$ $W_1$}\>\>\>\>\>(F$_2$)\>{\it $N_0$ \^etre Adj $W_2$ $W_1$}\\
(C)\>{\it $N_0$ Vsup $W_2$ un Napp $W_1$ Adj}\\
(C)\>{\it $N_0$ Vsup $W_2$ un Napp Adj $W_1$}\end{tabbing}
\caption{Adverbial complements of {\it Napp\/} and {\it Adj.} Free determiners are omitted.}
\label{avoirNappAdv}
\end{figure}

\section{Conclusion}

In fact, the prime motivation of this study was the description of adjectival kernel sentences.
Some adjectives seem to be intrinsically concerned with the construction (A) of
Fig.~\ref{avoirNapp}. This is the case of one of the two entries of {\it rapide\/} mentioned in
section~3.1. Either it occurs in relation with a {\it Vsup N\/} construction:
\begin{tabbing}(000)\=(D) =\=?*\=\kill
\>\>\>{\it Le convoi a une progression qui est rapide}\end{tabbing}
or with a noun phrase derived from that construction:
\begin{tabbing}(000)\=(D) =\=?*\=\kill
\>\>\>{\it La progression du convoi est rapide}\end{tabbing}
or such a construction is appropriate but can be reintroduced in the text through the
syntactic relations described above:
\begin{tabbing}(000)\=(D) =\=?*\=\kill
\>\>\>{\it Le convoi est rapide}\end{tabbing}
This analysis should facilitate the assignment of kernel sentences to such adjectives
and the description of lexical ambiguities.

Moreover, other types of predicates could have a similar property.
When {\it imiter\/} \cite[table 32R1]{BGL76} does not occur with a {\it Vsup N\/} construction:
\begin{tabbing}(000)\=(D) =\=?*\=\kill
\>\>\>{\it Luc fait des grimaces, Marie imite ces grimaces}\end{tabbing}
such a construction is implied, even though the sentence seems minimal:
\begin{tabbing}(000)\=(D) =\=?*\=\kill
\>\>\>{\it Marie imite Luc}\end{tabbing}
The complete description of kernel sentences for {\it rapide\/} and {\it imiter\/}
will probably depend on that of the nominal predicates, e.g. {\it progression\/} and
{\it grimace,} which are intrinsically associated to them.

\begin{figure}[h]
\caption{List of {\it Napp\/}'s with {\it Det Napp de Det $N_0$ \^etre Adj.}}\label{listeNapp}
\end{figure}

\begin{tabbing}
mode de fonctionn.\=austere, heterosexuel, libertin\=dente de, sinue de\=\kill
{\it Napp}\>{\it Adj}\>{\tt 12345678901234}\>{\sf Tables}\\
\\
{\it abn\a'egation}\>{\it h\a'ero\"{\i}que}\>{\tt +-++--+++-++++}\>{\sf APE1}\\
{\it abord}\>{\it engageant}\>{\tt +-++--+---++++}\>{\sf APP1 AN07}\\
{\it acad\a'emie}\>{\it beau}\>{\tt +-++--+----++-}\>{\sf APP1 AN07}\\
{\it accent}\>{\it lyrique}\>{\tt +++---++--+++-}\>{\sf APP1}\\
{\it acception}\>{\it p\a'ejoratif}\>{\tt -++---++--++++}\>{\sf APP1 AN07}\\
{\it acoustique}\>{\it mat, r\a'everb\a'erant}\>{\tt -++++-+---++++}\>{\sf APP1 AN07}\\
{\it acte}\>{\it odieux}\>{\tt +----+++--+++-}\>{\sf FN}\\
{\it activit\a'e}\>{\it f\a'ebrile}\>{\tt ++++--+---++++}\>{\sf AN01 AN10}\\
{\it adh\a'erence \a`a}\>{\it fort}\>{\tt -+++--+---++--}\>{\sf APP3}\\
{\it \^age}\>{\it ant\a'ediluvien}\>{\tt -+++--+-+-++-+}\>{\sf AN07}\\
{\it agissements}\>{\it louche}\>{\tt +-+--+-+--++-+}\\
{\it air}\>{\it bizarre}\>{\tt ++++--++--+++-}\>{\sf APP1 AN07}\\
{\it allure}\>{\it guind\a'e, louche}\>{\tt +-++--+++-++++}\>{\sf APP1 AN07}\\
{\it allure}\>{\it rapide}\>{\tt +++---+---++-+}\>{\sf APP1}\\
{\it altitude}\>{\it \a'elev\a'e}\>{\tt -++---+---++-+}\>{\sf APP1 AN07}\\
{\it ambiance}\>{\it sinistre, studieux}\>{\tt -++-+-+-+-++++}\>{\sf APP1 AN07}\\
{\it \^ame}\>{\it g\a'en\a'ereux}\>{\tt +-++--++----++}\>{\sf AN10}\\
{\it ampleur}\>{\it consid\a'erable}\>{\tt -+++--+-+-++-+}\>{\sf AN04}\\
{\it animation}\>{\it fourmillant}\>{\tt +++++-+-+-++++}\>{\sf APP1}\\
{\it aperture}\>{\it ferm\a'e, ouvert}\>{\tt -+++--+---++-+}\>{\sf AN07}\\
{\it apparat}\>{\it somptueux}\>{\tt -+++--+-+-++++}\>{\sf APP1 AN07}\\
{\it apparence}\>{\it glauque}\>{\tt ++++--++--++++}\>{\sf APP1 AN07}\\
{\it architecture}\>{\it baroque, classique}\>{\tt -+++--+---++++}\>{\sf AN07}\\
{\it ar\^ome}\>{\it parfum\a'e}\>{\tt -++---++++++++}\>{\sf AN05 AN07}\\
{\it ascendance}\>{\it m\a'etis}\>{\tt +-++--++------}\>{\sf AN10}\\
{\it aspect}\>{\it politique}\>{\tt -++---++-----+}\>{\sf APP1 AN08}\\
{\it aspect}\>{\it mat, brillant}\>{\tt ++++--++--++++}\>{\sf APP1 AN07}\\
{\it atmosph\a`ere}\>{\it sinistre, studieux}\>{\tt +++++-+---++++}\>{\sf AN07}\\
{\it attitude}\>{\it courtois, d\a'esinvolte, guind\a'e}\>{\tt +-++--++--++++}\>{\sf APP3 AN07}\\
{\it attrait}\>{\it irr\a'esistible}\>{\tt ++++--+++++++-}\>{\sf APP3 ES}\\
{\it bouquet}\>{\it parfum\a'e}\>{\tt -++---+-+-++++}\>{\sf APP1 AN07}\\
{\it brillant}\>{\it \a'eclatant}\>{\tt -+++--+---+++-}\>{\sf APP1 AN07}\\
{\it brio}\>{\it \a'eblouissant}\>{\tt +-++--+-+-++++}\>{\sf APP1 AN07}\\
{\it but}\>{\it professionnel}\>{\tt -++---++--++++}\\
{\it cadence}\>{\it effr\a'en\a'e}\>{\tt -++---++--+++-}\>{\sf APP1 AN07}\\
{\it calibre}\>{\it \a'enorme, monstrueux}\>{\tt -+++--+---+++-}\>{\sf AN07}\\
{\it capacit\a'e}\>{\it spacieux}\>{\tt -+++--+-+-++-+}\>{\sf AN07}\\
{\it caract\a`ere}\>{\it \a'epouvantable, souple}\>{\tt +-++--+-+-++++}\>{\sf APP1 AN07}\\
{\it caract\a'eristique}\>{\it v\a'eg\a'etal}\>{\tt +++---++--++++}\>{\sf APP1 AN08}\\
{\it charme}\>{\it irr\a'esistible, enchanteur}\>{\tt ++++--+-+-++++}\>{\sf APP3 ES AN07}\\
{\it classe}\>{\it international}\>{\tt ++++--+-+-++-+}\>{\sf APP1 AN07}\\
{\it climat}\>{\it \a'equatorial, temp\a'er\a'e}\>{\tt -++++-++--++-+}\>{\sf AN07}\\
{\it c\oe ur}\>{\it fid\a`ele, tendre}\>{\tt +-++--+-+---++}\>{\sf AN10}\\
{\it coloration}\>{\it clair, fonc\a'e}\>{\tt -+++--++--++-+}\>{\sf AN07}\\
{\it coloris}\>{\it pastel}\>{\tt -+++--++--++-+}\>{\sf AN07}\\
{\it combinatoire}\>{\it innombrable}\>{\tt -+++-++---++++}\>{\sf AN07}\\
{\it complexion}\>{\it d\a'elicat, robuste}\>{\tt +-++--+---++++}\>{\sf AN07}\\
{\it comportement}\>{\it gr\a'egaire, courtois}\>{\tt +++---++--++++}\>{\sf APP3 AN07}\\
{\it composante}\>{\it affectif}\>{\tt -++---++----++}\>{\sf AN08}\\
{\it composition}\>{\it m\a'elang\a'e}\>{\tt -+++--+---++-+}\\
{\it conception}\>{\it progressiste, traditionaliste}\>{\tt +-+----+--++++}\>{\sf APE3}\\
{\it condition}\>{\it bourgeois, roturier}\>{\tt +-++--+---++-+}\>{\sf AN07}\\
{\it conduite}\>{\it irr\a'eprochable, louche}\>{\tt +-+---+---++-+}\>{\sf AN07}\\
{\it configuration}\>{\it piriforme}\>{\tt -+++--+---++++}\>{\sf AN07}\\
{\it conformation}\>{\it longiligne}\>{\tt ++++--+---++++}\>{\sf AN07}\\
{\it connotation}\>{\it p\a'ejoratif}\>{\tt -++---++-+++++}\>{\sf APP1 AN07}\\
{\it consistance}\>{\it dur, mou}\>{\tt -+++--+--++++-}\>{\sf AN05}\\
{\it consonance}\>{\it harmonieux}\>{\tt -+++--++--++++}\>{\sf AN07}\\
{\it constitution}\>{\it d\a'elicat, robuste, athl\a'etique}\>{\tt ++++--+---++++}\>{\sf AN07}\\
{\it contact}\>{\it moelleux, soyeux}\>{\tt -+++--+---++++}\>{\sf AN07}\\
{\it contenance}\>{\it embarrass\a'e}\>{\tt +-+---+-+-+++-}\>{\sf APP1}\\
{\it contenance}\>{\it exigu, vaste}\>{\tt -+++--+---++-+}\>{\sf APP1 AN07}\\
{\it contenu}\>{\it philosophique}\>{\tt -+++--+---++++}\>{\sf AN08}\\
{\it contour}\>{\it carr\a'e, polygonal, rond}\>{\tt -++---++--++--}\>{\sf AN07}\\
{\it cote}\>{\it bon march\a'e, ruineux}\>{\tt -++---+---++--}\>{\sf APE3 DR1 AN07}\\
{\it c\^ot\a'e}\>{\it polisson}\>{\tt +++---++----++}\>{\sf APP1 AN08 AN10}\\
{\it couleur}\>{\it beige}\>{\tt -+++--++++++--}\>{\sf AN06 APP1}\\
{\it coupe}\>{\it \a'etriqu\a'e}\>{\tt -+++--+---+++-}\>{\sf DR2 AN07}\\
{\it courant}\>{\it torrentiel}\>{\tt -++++-+-+-++++}\>{\sf AN08}\\
{\it cours}\>{\it sinueux}\>{\tt -++---+---++-+}\>{\sf AN09}\\
{\it cours}\>{\it bon march\a'e, ruineux}\>{\tt -++---++--++-+}\>{\sf APP1 AN09}\\
{\it course}\>{\it rapide}\>{\tt +++---+----+-+}\\
{\it co\^ut}\>{\it bon march\a'e, ruineux}\>{\tt -+++--++--++-+}\>{\sf AN04 AN07}\\
{\it cubage}\>{\it exigu, spacieux}\>{\tt -+++--+---++-+}\>{\sf F2A AN07}\\
{\it datation par}\>{\it ant\a'ediluvien}\>{\tt +++---+---++--}\>{\sf FR1 AN07}\\
{\it date}\>{\it ancien, ant\a'erieur \a`a de}\>{\tt -+++--+---++--}\>{\sf AN07}\\
{\it d\a'ebit}\>{\it torrentiel}\>{\tt -++++-+-+-++++}\>{\sf APP1 AN07}\\
{\it d\a'ecor}\>{\it damasquin\a'e}\>{\tt -+++--+---++-+}\>{\sf APP1 AN07}\\
{\it d\a'egaine}\>{\it photog\a'enique}\>{\tt +-++--+---++++}\>{\sf APP1 AN07}\\
{\it dehors}\>{\it avenant, n\a'eglig\a'e}\>{\tt +-+----+---+++}\\
{\it d\a'emarche}\>{\it boitillant}\>{\tt +-++--+---++++}\>{\sf AN10}\\
{\it dessin}\>{\it sinueux}\>{\tt -+++--+---++++}\>{\sf AN07}\\
{\it d\a'etermination \a`a}\>{\it inflexible}\>{\tt +-++--+-+-++++}\>{\sf AN02 AN07}\\
{\it diction}\>{\it bredouillant}\>{\tt +-+---+---++++}\>{\sf APP1}\\
{\it dimension}\>{\it \a'enorme, minuscule}\>{\tt -+++-+-+--++++}\>{\sf APP1 AN07}\\
{\it direction}\>{\it parall\a`ele \a`a, perpendiculaire \a`a}\>{\tt -++---+---++-+}\>{\sf AN09}\\
{\it disposition}\>{\it ordonn\a'e}\>{\tt -++---+---++++}\\
{\it disposition}\>{\it jaloux, tourment\a'e}\>{\tt +-++--++--++++}\\
{\it dominante}\>{\it lyrique}\>{\tt -+++--++--++++}\>{\sf AN08}\\
{\it dur\a'ee}\>{\it bref}\>{\tt -+++--+---++-+}\>{\sf AN07}\\
{\it \a'echelle}\>{\it r\a'eduit}\>{\tt -+++--+---++--}\>{\sf AN09}\\
{\it \a'eclat}\>{\it \a'eblouissant}\>{\tt -+++--+++-++++}\>{\sf APP1 AN07}\\
{\it \a'economie}\>{\it structur\a'e Advm}\>{\tt -+++--+---++++}\>{\sf AN07}\\
{\it \a'economie}\>{\it mis\a'erable, prosp\a`ere}\>{\tt -+++--+---++++}\>{\sf AN08}\\
{\it \a'ecriture }\>{\it illisible}\>{\tt -++---+---++++}\>{\sf APP1 AN08}\\
{\it \a'education}\>{\it raffin\a'e}\>{\tt +-++--+-+-++++}\>{\sf APE1 DR1 AN07}\\
{\it effectif}\>{\it pl\a'ethorique}\>{\tt -+++--++--++-+}\>{\sf AN07}\\
{\it effluve}\>{\it aromatique, naus\a'eabond}\>{\tt -++----+--++++}\>{\sf AN09}\\
{\it \a'elocution}\>{\it bredouillant}\>{\tt +-+---+---++++}\>{\sf APP1 AN10}\\
{\it emploi}\>{\it p\a'ejoratif}\>{\tt -+++--++--++++}\>{\sf AN07}\\
{\it empressement \a`a}\>{\it diligent}\>{\tt +-++--+-+-++++}\>{\sf APE3}\\
{\it \a'energie}\>{\it in\a'epuisable}\>{\tt +-++--+-+-++++}\>{\sf AN03}\\
{\it entrain}\>{\it joyeux}\>{\tt +++++-+-+-++++}\>{\sf APP1 AN07}\\
{\it entrain \a`a}\>{\it diligent}\>{\tt +-+---+-+-++++}\>{\sf APP3}\\
{\it envergure}\>{\it consid\a'erable, limit\a'e}\>{\tt -+++--+-+-++++}\>{\sf APP1 AN07}\\
{\it esprit}\>{\it fin, subtil}\>{\tt +-++--+-+-++++}\>{\sf APP1 AN07 AN10}\\
{\it essence}\>{\it sup\a'erieur, transcendant}\>{\tt ++-+--+---++++}\>{\sf AN07}\\
{\it esth\a'etique}\>{\it romantique}\>{\tt -++++-+---++++}\>{\sf AN07}\\
{\it \a'etat}\>{\it catastrophique, satisfaisant}\>{\tt -+----+---+++-}\>{\sf AN09}\\
{\it \a'etendue}\>{\it vaste}\>{\tt -+++--+-+-++-+}\>{\sf AN07}\\
{\it \a'etendue}\>{\it consid\a'erable, limit\a'e, vaste}\>{\tt -+++--+---++-+}\>{\sf AN07}\\
{\it existence}\>{\it routinier}\>{\tt +-+---+---++-+}\>{\sf AN10}\\
{\it expression}\>{\it grima\c{c}ant, souriant}\>{\tt ++++-++++-++++}\>{\sf APP1 AN10}\\
{\it extension}\>{\it local, national, plan\a'etaire}\>{\tt -+++--+-+-++++}\>{\sf APP3 AN07}\\
{\it ext\a'erieur}\>{\it avenant, engageant, n\a'eglig\a'e}\>{\tt ++++--+---++++}\>{\sf AN07}\\
{\it fa\c{c}on de}\>{\it affect\a'e}\>{\tt +++---++--++++}\\
{\it fa\c{c}on}\>{\it familier}\>{\tt +-+----+--++++}\>{\sf AN10}\\
{\it faconde}\>{\it prolixe}\>{\tt +-++--+++-+++-}\>{\sf APP1 AN10}\\
{\it facture}\>{\it dada\"{\i}ste}\>{\tt -+++--+---++++}\>{\sf AN07}\\
{\it faste}\>{\it somptueux}\>{\tt -+++--+-+-++++}\>{\sf AN03}\\
{\it finalit\a'e}\>{\it professionnel}\>{\tt -++---+---++++}\>{\sf AN08}\\
{\it fond}\>{\it bienveillant}\>{\tt +-++--+-----++}\>{\sf AN10}\\
{\it force}\>{\it in\a'epuisable}\>{\tt +-+----+--+++-}\>{\sf APE11 AN10}\\
{\it format}\>{\it encombrant}\>{\tt -+++--+---++++}\>{\sf APP1 AN07}\\
{\it forme}\>{\it allong\a'e, arrondi, carr\a'e}\>{\tt -+++--++--++++}\>{\sf APP1 AN07}\\
{\it forme}\>{\it plantureux}\>{\tt +-++---+--++++}\>{\sf APP1 AN10}\\
{\it fr\a'equence}\>{\it incessant, rapproch\a'e}\>{\tt -+++--+---++-+}\>{\sf AN05 AN07}\\
{\it fr\a'equentation}\>{\it recommandable}\>{\tt +--+--+-------}\\
{\it geste}\>{\it adroit, g\a'en\a'ereux, magnanime}\>{\tt +-+--+++--++++}\>{\sf FN}\\
{\it go\^ut}\>{\it sobre}\>{\tt +-++--+++-++++}\>{\sf APP1}\\
{\it go\^ut}\>{\it acide, amer, douce\^atre}\>{\tt -+++--+-+-++++}\>{\sf APP1 AN07}\\
{\it grain}\>{\it fin, grossier}\>{\tt -+++--+---++++}\>{\sf AN06 AN07}\\
{\it graphisme}\>{\it \a'el\a'egant}\>{\tt -+++--+---++++}\>{\sf AN07}\\
{\it habitat}\>{\it cavernicole}\>{\tt -++---+---++--}\>{\sf AN08}\\
{\it heure}\>{\it tardif}\>{\tt -++---+---++--}\\
{\it humeur}\>{\it enjou\a'e, morose}\>{\tt +-++--+---++++}\>{\sf AN07}\\
{\it imagination}\>{\it cr\a'eatif}\>{\tt +-++--+-+-++++}\>{\sf AN02 AN10}\\
{\it inconfort}\>{\it spartiate}\>{\tt -++++-+---++++}\>{\sf AN04 AN07}\\
{\it inspiration}\>{\it dada\"{\i}ste, mystique}\>{\tt -+++--+---++++}\>{\sf APE1 AN07}\\
{\it instinct}\>{\it maternel}\>{\tt +-+---++++++++}\>{\sf AN07}\\
{\it issue}\>{\it fatal}\>{\tt -++---+---++-+}\>{\sf AN07}\\
{\it langage}\>{\it ordurier, po\a'etique}\>{\tt +-+---+---++++}\>{\sf APP1 AN10}\\
{\it lustre}\>{\it \a'eblouissant, \a'eclatant}\>{\tt -+++--+-+-++++}\>{\sf APP1 AN07}\\
{\it maintien}\>{\it guind\a'e}\>{\tt +-++--+---++++}\>{\sf AN10}\\
{\it mani\a`ere de}\>{\it affect\a'e}\>{\tt +++---++--++++}\>{\sf APP3}\\
{\it mani\a`ere}\>{\it affect\a'e}\>{\tt +-+----+--++++}\>{\sf APP3 AN10}\\
{\it marque}\>{\it cor\a'een, fran\c{c}ais}\>{\tt -+++--+---++-+}\>{\sf AN07}\\
{\it mati\a`ere}\>{\it compact}\>{\tt -+++--+---++++}\>{\sf AN07}\\
{\it mentalit\a'e}\>{\it infantile}\>{\tt +-++--+---++++}\>{\sf AN07}\\
{\it mine}\>{\it rayonnant de}\>{\tt +-++-+++--+++-}\>{\sf APP1 AN07 AN10}\\
{\it mise}\>{\it d\a'ebraill\a'e}\>{\tt +-++--++--++++}\>{\sf AN07}\\
{\it mode d'emploi}\>{\it commode}\>{\tt -+++--+---++++}\\
{\it mode de fonctionn.}\>{\it automatique}\>{\tt -++---+---++-+}\\
{\it m\oe urs}\>{\it aust\a`ere, h\a'et\a'erosexuel, libertin}\>{\tt +-++---+--++-+}\>{\sf APP1 AN07}\\
{\it montant}\>{\it \a'elev\a'e, minime}\>{\tt -+++--+---++-+}\>{\sf AN07}\\
{\it morale}\>{\it aust\a`ere, rel\^ach\a'e}\>{\tt +-++--+---++++}\>{\sf AN07 AN10}\\
{\it morgue}\>{\it hautain}\>{\tt +-++--+-+-++++}\>{\sf APP3}\\
{\it nature}\>{\it adipeux}\>{\tt ++++--+---++++}\>{\sf AN07}\\
{\it niveau}\>{\it avanc\a'e}\>{\tt ++++--++--++-+}\>{\sf AN07}\\
{\it nombre}\>{\it pl\a'ethorique}\>{\tt ++----+---++-+}\>{\sf AN08}\\
{\it notori\a'et\a'e}\>{\it confidentiel}\>{\tt -+++--+-+-+++-}\>{\sf AN04 APP3}\\
{\it nuance}\>{\it orang\a'e}\>{\tt -+++--++-+++--}\>{\sf APP1 AN07}\\
{\it ob\a'edience}\>{\it catholique, communiste}\>{\tt +-++--++--++-+}\>{\sf AN07}\\
{\it objectif}\>{\it professionnel}\>{\tt -++---++--++++}\\
{\it organisation}\>{\it pentam\a`ere, rayonn\a'e}\>{\tt -++---+---++-+}\>{\sf APE1 AN07}\\
{\it origine}\>{\it autochtone, indig\a`ene}\>{\tt -+++--++--++++}\>{\sf AN07}\\
{\it parole}\>{\it optimiste, pessimiste}\>{\tt +-+----+--++-+}\\
{\it partie}\>{\it apocryphe, authentique}\>{\tt -++---++----++}\>{\sf AN08}\\
{\it pens\a'ee}\>{\it rationnel}\>{\tt +-+---+---++++}\>{\sf APE3 AN10}\\
{\it pente}\>{\it abrupt}\>{\tt -+++--+---++-+}\>{\sf AN08}\\
{\it personnalit\a'e}\>{\it haut en couleur}\>{\tt +-++--+-+-++++}\>{\sf AN07}\\
{\it pesanteur}\>{\it \a'ecrasant}\>{\tt -+++--+---++++}\>{\sf APP1 AN07}\\
{\it pH}\>{\it acide, amphot\a`ere, basique}\>{\tt -+++--+---++-+}\\
{\it phase}\>{\it manuelle}\>{\tt +++---++-----+}\>{\sf AN08}\\
{\it physionomie}\>{\it renfrogn\a'e, sombre}\>{\tt +-++--+---++++}\>{\sf AN07}\\
{\it physique}\>{\it corpulent}\>{\tt +-++--+---++++}\>{\sf APP1 AN07}\\
{\it plan}\>{\it en damier, g\a'eom\a'etrique}\>{\tt -++---+---++-+}\\
{\it plastique}\>{\it sculptural}\>{\tt +-++--+---++++}\>{\sf APP1 AN07}\\
{\it plume}\>{\it tourment\a'e}\>{\tt +-+---+----++-}\>{\sf AN10}\\
{\it poids}\>{\it \a'ecrasant}\>{\tt -+++--+---++++}\>{\sf AN07}\\
{\it poids}\>{\it capital, d\a'ecisif}\>{\tt -+++--+-+-++++}\>{\sf APP1 APP3 AN07}\\
{\it port\a'ee}\>{\it historique}\>{\tt -+++--+-+-++++}\>{\sf APP1 AN07}\\
{\it position}\>{\it universaliste}\>{\tt +-++--++--++-+}\>{\sf APP2 AN10}\\
{\it posture}\>{\it accroupi}\>{\tt +-+---++--++--}\>{\sf APP1 AN10}\\
{\it pr\a'esence}\>{\it imposant}\>{\tt +-++--+-+-++++}\>{\sf AN01 AN07}\\
{\it prestance}\>{\it imposant}\>{\tt ++++--+-+-++++}\>{\sf APP1 AN07}\\
{\it prix}\>{\it inestimable}\>{\tt ++++--+-+-+++-}\>{\sf AN05 APP3 AN07}\\
{\it profil}\>{\it galb\a'e}\>{\tt +++---+---++++}\>{\sf AN10}\\
{\it progression Loc}\>{\it lent, rapide}\>{\tt +++---+---++-+}\>{\sf F1C}\\
{\it proportion}\>{\it d\a'emesur\a'e}\>{\tt +++----+-+++++}\\
{\it propos \a`a}\>{\it grandiloquent, insipide}\>{\tt +-+----+--++-+}\\
{\it psychisme}\>{\it infantile, parano\"{\i}aque}\>{\tt +-++--+---++++}\>{\sf AN10}\\
{\it qualit\a'e}\>{\it excellent}\>{\tt -+-+--+---++-+}\>{\sf AN07}\\
{\it qualit\a'e}\>{\it m\a'eritant}\>{\tt +-+---++-+++++}\>{\sf APP1}\\
{\it quantit\a'e}\>{\it abondant}\>{\tt ++----++--++-+}\\
{\it race}\>{\it b\^atard, crois\a'e, m\a'etiss\a'e}\>{\tt +-++--+---++-+}\>{\sf AN07}\\
{\it raisonnement}\>{\it tortueux}\>{\tt +-+---++--++++}\>{\sf F8}\\
{\it rang Loc}\>{\it \a'eminent, subalterne}\>{\tt +-++--+---++-+}\>{\sf APP3 AN07}\\
{\it rectitude}\>{\it irr\a'eprochable}\>{\tt +-++--+-+-++++}\>{\sf APP1 AN07}\\
{\it r\a'egime}\>{\it carnivore, herbivore}\>{\tt +-+---+---++-+}\\
{\it r\a'egime}\>{\it r\a'egulier, irr\a'egulier}\>{\tt -++---+---++++}\\
{\it registre}\>{\it mystique, tragique}\>{\tt -+-+--++--++++}\>{\sf APP1 AN07}\\
{\it relent}\>{\it malodorant}\>{\tt -++---++--++++}\>{\sf APP1}\\
{\it relief}\>{\it montagneux}\>{\tt -++---++++++-+}\>{\sf APP1 AN08}\\
{\it rendement}\>{\it lucratif}\>{\tt -+++--+++-++++}\>{\sf APP1 AN07}\\
{\it renom}\>{\it illustre}\>{\tt ++++--+---++++}\>{\sf AN01 AN07}\\
{\it renomm\a'ee}\>{\it illustre}\>{\tt ++++--+---++++}\>{\sf AN01 AN07}\\
{\it r\a'eputation}\>{\it illustre}\>{\tt ++++--+---++++}\>{\sf AN01 APP3 AN07}\\
{\it r\a'esonance}\>{\it nasillard}\>{\tt -+++--+---++++}\>{\sf APP3 AN07}\\
{\it rh\a'etorique}\>{\it grandiloquent, pompeux}\>{\tt -+++--+---++++}\>{\sf AN07}\\
{\it rythme}\>{\it effr\a'en\a'e, hach\a'e}\>{\tt -+++--+++-++++}\>{\sf AN07}\\
{\it sang}\>{\it noble}\>{\tt +-++--+----+++}\>{\sf AN07}\\
{\it saveur}\>{\it relev\a'e}\>{\tt -+++--+++-+++-}\>{\sf AN05 APP3 AN07}\\
{\it savoir}\>{\it \a'erudit}\>{\tt +-+---+---+++-}\>{\sf APE3 AN07}\\
{\it section}\>{\it carr\a'e}\>{\tt -++---+---++--}\\
{\it s\a'eduction}\>{\it irr\a'esistible}\>{\tt ++++--+-+-++++}\>{\sf APP3 ES}\\
{\it s\a'emantique}\>{\it clair, obscur}\>{\tt -+++--+---++++}\>{\sf AN07}\\
{\it sens}\>{\it politique, tactique}\>{\tt -++---++--+++-}\>{\sf APP3 AN08}\\
{\it sensation}\>{\it euphorique}\>{\tt +-+---++--+++-}\>{\sf APE3}\\
{\it senteur}\>{\it parfum\a'e}\>{\tt -+++--++--+++-}\>{\sf APP1 AN07}\\
{\it sentiment}\>{\it joyeux, nostalgique}\>{\tt +-+---++-++++-}\>{\sf APE3}\\
{\it s\a'erieux}\>{\it scrupuleux}\>{\tt +-++--+-+-++++}\>{\sf AN03 AN07}\\
{\it sexe}\>{\it f\a'eminin, masculin}\>{\tt ++++--++--++--}\>{\sf AN06 AN07}\\
{\it silhouette}\>{\it galb\a'e}\>{\tt +++---+---++++}\>{\sf APP1 AN10}\\
{\it son}\>{\it clair, nasal}\>{\tt -++---+---++++}\>{\sf AN07}\\
{\it sonorit\a'e}\>{\it clair, nasal}\>{\tt -+++--+---++++}\>{\sf AN05 AN07}\\
{\it souffle}\>{\it \a'epique}\>{\tt -++++-+-+-++++}\>{\sf APP1 AN07}\\
{\it standing}\>{\it opulent}\>{\tt -+++--+-+-+++-}\>{\sf APP1 AN07}\\
{\it standing}\>{\it ais\a'e}\>{\tt +-+---+-+-+++-}\>{\sf APP1 AN10}\\
{\it stature}\>{\it imposant}\>{\tt ++++--+---++++}\>{\sf AN07}\\
{\it statut}\>{\it officiel}\>{\tt -++---+---++++}\>{\sf APP3 DR2}\\
{\it structure}\>{\it ramifi\a'e, rayonnant}\>{\tt -+++--+---++-+}\>{\sf AN07}\\
{\it style}\>{\it archa\"{\i}sant, pompeux}\>{\tt ++++--+-+-++++}\>{\sf APP1 AN07}\\
{\it substance}\>{\it fluide, visqueux}\>{\tt -+++--+---++++}\>{\sf AN07}\\
{\it substance}\>{\it dense}\>{\tt -+++--+-+-++++}\>{\sf AN07}\\
{\it superficie}\>{\it \a'etendu, vaste}\>{\tt -+++--+-+-++-+}\\
{\it surface}\>{\it lisse, rugueux}\>{\tt -+++--+-+-++++}\>{\sf AN07}\\
{\it taille}\>{\it grand, petit}\>{\tt ++++--+---++-+}\>{\sf AN07}\\
{\it tarif}\>{\it ruineux}\>{\tt -++---++--++-+}\>{\sf AN07}\\
{\it taux}\>{\it ruineux}\>{\tt -++---++--++--}\>{\sf AN07}\\
{\it teint}\>{\it bronz\a'e, livide}\>{\tt +-++--+---++++}\>{\sf AN07}\\
{\it teinte}\>{\it chamois}\>{\tt -+++--++--++-+}\>{\sf AN07}\\
{\it temp\a'erament}\>{\it optimiste, pessimiste}\>{\tt +-++--+-+-++++}\>{\sf APP1 AN07}\\
{\it temp\a'erature}\>{\it br\^ulant}\>{\tt -++-+-+---++-+}\>{\sf AN07}\\
{\it tempo}\>{\it allegro}\>{\tt -+++--++--++-+}\>{\sf APP1 AN07}\\
{\it tenue}\>{\it d\a'ebraill\a'e}\>{\tt +-+---++--++++}\>{\sf AN07}\\
{\it texture}\>{\it l\^ache, serr\a'e}\>{\tt -+++--+---++++}\>{\sf APP1 AN07}\\
{\it timbre}\>{\it clair, nasal, sombre}\>{\tt ++++--+-+-++++}\>{\sf APP1 AN07}\\
{\it ton}\>{\it chamois}\>{\tt -+++--++--++-+}\>{\sf AN07}\\
{\it ton}\>{\it doucereux, p\a'edant}\>{\tt +++---+---++++}\>{\sf APP3 AN07}\\
{\it tonalit\a'e}\>{\it chantant}\>{\tt -+++--++--++++}\>{\sf AN07}\\
{\it toucher}\>{\it moelleux, soyeux}\>{\tt -+++--+---++++}\\
{\it tournure}\>{\it encourageant}\>{\tt -++---+---+++-}\>{\sf APP1 AN07}\\
{\it trac\a'e}\>{\it sinueux}\>{\tt -+++--+---++-+}\>{\sf AN07}\\
{\it trait}\>{\it totalitaire}\>{\tt -++---++----++}\>{\sf APP1 AN08}\\
{\it trait}\>{\it grima\c{c}ant, souriant}\>{\tt ++++---+--++++}\>{\sf APP1 AN10}\\
{\it tranchant}\>{\it \a'emouss\a'e}\>{\tt -+++--+---++--}\>{\sf AN07}\\
{\it type}\>{\it am\a'erindien}\>{\tt +-++--+---++++}\>{\sf AN07}\\
{\it usage aupr\a`es de}\>{\it commode, courant, prohib\a'e}\>{\tt -+++--+---++-+}\>{\sf AN07}\\
{\it valeur}\>{\it in\a'egal, inestimable}\>{\tt -+++--+-+-++++}\>{\sf APP3 AN07}\\
{\it v\a'eg\a'etation}\>{\it luxuriant, souffreteux}\>{\tt -++++-+---++++}\>{\sf AN08}\\
{\it vie}\>{\it heureux}\>{\tt +-+---+---++-+}\>{\sf AN10}\\
{\it voix}\>{\it tonitruant}\>{\tt +++---+-+-+++-}\>{\sf APP1 AN10}\\
{\it volont\a'e de}\>{\it inflexible}\>{\tt +-++--+-+-++++}\>{\sf APE3 AN07}\\
{\it volume}\>{\it \a'enorme}\>{\tt -+++-++-+-++++}\>{\sf AN05 APP1 AN07}\\
{\it vue}\>{\it universaliste}\>{\tt +-+----+--++-+}\>{\sf AN08}\\
{\it z\a`ele \a`a}\>{\it empress\a'e}\>{\tt +-++--+-+-++++}\>{\sf AN02 APE3}\end{tabbing}

\end{document}